\documentclass{article}

\PassOptionsToPackage{numbers}{natbib}
\usepackage[main,final]{neurips_2026}

\PassOptionsToPackage{table}{xcolor}

\usepackage[dvipsnames,table]{xcolor}
\definecolor{myblue}{rgb}{.894,.937,.965}
\definecolor{mydark_blue}{rgb}{.68,.776,.906}
\definecolor{NO3}{rgb}{.749,.902,.808}
\definecolor{NO2}{rgb}{.999,.973,.773}
\definecolor{NO1}{rgb}{0.98, 0.78, 0.57}
\usepackage[utf8]{inputenc}
\usepackage[T1]{fontenc}
\usepackage{hyperref}
\usepackage{url}
\usepackage{booktabs}
\usepackage{amsfonts}
\usepackage{nicefrac}
\usepackage{microtype}
\usepackage{xcolor}
\usepackage{mathtools}
\usepackage{caption}
\usepackage{xspace}
\usepackage{multirow}
\usepackage{pifont}
\usepackage{wrapfig}
\usepackage{tabularx}
\usepackage{placeins}
\usepackage{needspace}

\hypersetup{hidelinks,
  pdftitle={UNIVERSE: Unified Video Action Models for Autonomous Driving with Flexible Mask-Modulated Modality Generation},
  pdfauthor={Mengmeng Liu, Diankun Zhang, Jiuming Liu, Jianfeng Cui, Hongwei Xie, Guang Chen, Hangjun Ye, Francesco Nex, Hao Cheng, Michael Ying Yang}}

\newcommand{\ourmethod}{\textsc{UNIVERSE}\xspace}

\definecolor{gtfuture}{RGB}{44,160,44}
\definecolor{predfuture}{RGB}{214,39,40}
\definecolor{dpvogat}{RGB}{37,99,235}
\definecolor{dpvopred}{RGB}{245,158,11}
\usepackage{enumitem}
\usepackage{graphicx}
\usepackage{booktabs}
\usepackage{amssymb}

\newcommand{\cmark}{\textcolor{green!70!black}{\ding{51}}} % green check
\newcommand{\xmark}{\textcolor{red!70!black}{\ding{55}}}   % red cross 
\title{UNIVERSE: Unified Video Action Models \\for Autonomous Driving with Flexible \\ Mask-Modulated Modality Generation}

\author{%
  \textbf{Mengmeng Liu}\textsuperscript{1}
  \quad \textbf{Diankun Zhang}\textsuperscript{2}
  \quad \textbf{Jiuming Liu}\textsuperscript{3}
  \quad \textbf{Jianfeng Cui}\textsuperscript{2}
  \\[0.3em]
  \textbf{Hongwei Xie}\textsuperscript{2}
  \quad \textbf{Guang Chen}\textsuperscript{2}
  \quad \textbf{Hangjun Ye}\textsuperscript{2}
  \\[0.3em]
  \textbf{Francesco Nex}\textsuperscript{1}
  \quad \textbf{Hao Cheng}\textsuperscript{1}
  \quad \textbf{Michael Ying Yang}\textsuperscript{4,*}
  \\[0.7em]
  \normalfont\textsuperscript{1}University of Twente
  \quad \textsuperscript{2}Xiaomi EV
  \\[0.2em]
  \normalfont\textsuperscript{3}University of Cambridge
  \quad \textsuperscript{4}University of Bath
  \\[0.3em]
  \normalfont\textsuperscript{*}Corresponding author.
}

\begin{document}

\maketitle

\begin{abstract}
World Action Models (WAMs) have shown strong potential for improving action generalization in autonomous driving by using future video prediction as dense supervision for scene dynamics. However, retaining this supervision without requiring expensive future-video generation at deployment remains an architectural challenge. Existing designs either separate video and action parameters or couple their future-token dependencies during joint denoising. We propose \textsc{UNIVERSE}, a unified video-action model built upon a \emph{single mask-modulated Diffusion Transformer}. It retains shared video-action parameters so that dense video supervision directly shapes trajectory denoising, while introducing a \emph{Modality-Decoupling Visibility Mask} that shares historical context and blocks mutual attention between future video and trajectory tokens. This makes future-video generation optional: removing future-video tokens at test time preserves the trajectory branch's training-time attention dependencies, enabling trajectory-only inference with a $4.3\times$ speedup over coupled video-action rollout and comparable planning accuracy. The same model also supports video-only and joint video-action rollouts. Experiments show that \textsc{UNIVERSE} achieves 91.0 PDMS on NAVSIM (vs. 89.6 for the Two-DiT variant) and strong zero-shot transfer to nuScenes and Bench2Drive without fine-tuning. Controlled comparisons and ablations on pretraining and loss weighting support the roles of shared parameters, video co-training, and mask-based modality decoupling.
\par\smallskip
\noindent\textbf{Code:} \url{https://github.com/xiaomi-mlab/UNIVERSE}
\end{abstract}

\section{Introduction}

Generalization to unseen conditions is a central challenge in visual learning~\cite{tang2026learning,liu20264dstr,lu2026worldcoder,liu2026arflow,lu2025uni,dong2026language,dong2026towards,liu2026palm,shen2026fine,yu2026elsa3d,dong2025ha,meng2026phymetric} and autonomous driving~\cite{liu2026streamvlo,liu2025dvlo4d,liu2023tracing,zhang2026diffusion,Deng_2026_CVPR,deng2026gaussiandwm++,liu2024laformer,cheng2023gatraj,liu2023regformer,liu2025difflow3d,xie2026step}, where a planner trained on one dataset must safely transfer to unseen cities, traffic patterns, sensor distributions, and simulation environments.
Standard end-to-end planners directly map historical observations to future ego trajectories, which is efficient but can overfit dataset-specific driving priors such as common trajectory patterns, road layouts, or benchmark-specific behaviors~\cite{hu2023planning,chitta2022transfuser,liao2025diffusiondrive}.
World Action Models (WAMs) offer a promising alternative by coupling future prediction with action generation~\cite{ye2026world}.
Compared with sparse trajectory supervision, future world modeling provides dense signals about scene geometry, object motion, traffic-agent interactions, and temporal causality, helping the model learn both \emph{how the world evolves} and \emph{how the ego vehicle should act}~\cite{lecun2022path,ha2018world,wang2024driving,zheng2024genad}.
Recent driving world models and embodied video-action models further show that future visual prediction can improve planning, simulation, transferability, and zero-shot policy generalization~\cite{gao2024vista,ye2026world,jang2025dreamgen,lingbot-va2026,li2025unified,zhu2025unified,lu2026one}.

A key question is whether these training benefits require expensive future-video generation at deployment.
DriveVA~\cite{liu2026driveva}, the direct predecessor of this work, establishes the feasibility of zero-shot driving through joint video-trajectory denoising in a shared DiT; its future video and trajectory tokens interact directly.
Fast-WAM~\cite{yuan2026fast} instead shows that future-video prediction can serve as training-time supervision while action inference bypasses video rollout.
These findings motivate a complementary question: \emph{how can video supervision train the same parameters used for trajectory generation while future-video generation remains optional?}
Cascaded or dual-DiT designs separate video and action parameters~\cite{li2025imagidrive,yuan2026fast,ye2026gigaworldpolicy,wang2025prophetdwm}, whereas a coupled shared-DiT design requires future-video tokens in the trajectory computation graph.
Simply deleting those tokens at deployment changes the dependencies learned during training.

\begin{figure}[t]
    \centering
    \includegraphics[width=0.98\linewidth]{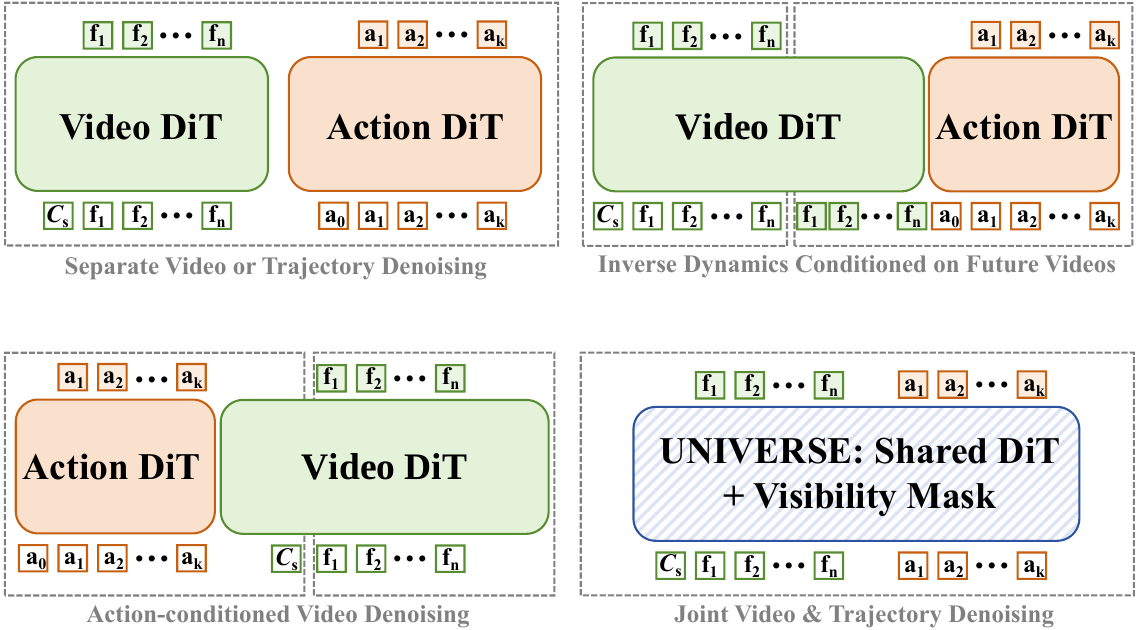}
    \caption{\textbf{Video-action denoising dependency patterns.}
    $C_s$ denotes clean history and current ego-state conditioning; $f_i$ and $a_i$ denote video-frame and trajectory tokens.
    The first three panels illustrate separate or conditionally coupled modules; the last shows \textsc{UNIVERSE}'s shared DiT with future-modality visibility decoupling.}
    \label{fig:comparison}
\end{figure}

We propose \textsc{UNIVERSE}, a unified video-action model built upon a \emph{single mask-modulated Diffusion Transformer}~\cite{peebles2023DIT}.
It retains the shared video-action backbone so that both losses update the same attention and feed-forward parameters, rather than regularizing a separate trajectory model indirectly.
Our new design concerns the \emph{future-token dependency structure}, not the use of a pretrained video backbone or shared DiT alone.
The \emph{Modality-Decoupling Visibility Mask} allows both future modalities to read common historical context, blocks mutual attention between future video and trajectory tokens, and prevents context tokens from relaying information from either future modality.
This yields \emph{coupling-by-parameters} and \emph{decoupling-by-visibility}: training retains dense video supervision, while either future modality can be instantiated independently at inference.

Trajectory-only inference therefore removes future-video tokens without changing the trajectory branch's attention dependencies, reducing latency by $4.3\times$ relative to coupled rollout.
The same weights support video-only and joint outputs for future-scene visualization and video-trajectory consistency analysis.
Controlled no-video, no-mask, and Two-DiT experiments distinguish the roles of supervision, visibility, and parameter sharing.
The cross-domain results support improved transfer; they do not by themselves establish that source-domain action priors are the sole cause of generalization errors.

\Needspace{5\baselineskip}
Our contributions are:
\begin{itemize}
    \item A modality-decoupled shared-DiT formulation that preserves video supervision of trajectory-generation parameters while making future-video generation optional at deployment.
    \item A parameter-free Modality-Decoupling Visibility Mask and output-dependent token instantiation that support dependency-consistent trajectory-only, video-only, and joint inference with the same model weights.
    \item An empirical study of planning, cross-domain transfer, video-action consistency, and computational cost, including comparisons with close prior methods and controlled variants. \textsc{UNIVERSE} achieves 91.0 NAVSIM PDMS, strong zero-shot transfer, and $4.3\times$ faster trajectory inference than coupled rollout.
\end{itemize}

\section{Related Work}
\label{sec:related_w}
\subsection{World Models for Autonomous Driving Video Generation}

World modeling~\cite{chen2026memobench,shi2026honeycomb,liu2026towards} has become a major direction for autonomous-driving video generation, aiming to synthesize temporally coherent and controllable future observations for simulation, data augmentation, and downstream decision-making. 
Early works such as DriveGAN~\cite{kim2021drivegan} and MagicDrive~\cite{gao2023magicdrive} explore controllable driving-scene generation. 
Recent methods further improve multi-view consistency, 3D geometric control, long-horizon generation, and scenario diversity. 
Representative examples include DriveDreamer~\cite{wang2024drivedreamer}, DriveDreamer-2~\cite{zhao2025drivedreamer}, Panacea~\cite{wen2024panacea}, VISTA~\cite{gao2024vista}, GAIA-1~\cite{hu2023gaia}, GAIA-2~\cite{russell2025gaia}, MagicDrive-V2~\cite{gao2025magicdrive}, MiLA~\cite{wang2025mila}, CoGen~\cite{ji2025cogen}, Genesis~\cite{guo2025genesis}, and OmniNWM~\cite{li2025omninwm}. 
Another line of work studies world models in BEV, occupancy, LiDAR, or 4D representations, such as OccWorld~\cite{zheng2024occworld}, DriveWorld~\cite{min2024driveworld}, DrivingWorld~\cite{hu2024drivingworld}, WoVoGen~\cite{lu2024wovogen}, UniScene~\cite{li2025uniscene}, and related occupancy-centric driving simulators.
These methods have significantly advanced the fidelity, controllability, and temporal consistency of driving-scene generation. 
However, they are primarily designed as scene simulators or data generators, with the main objective of improving visual realism, geometric consistency, or controllable rollout. 
Less attention has been paid to how video-learned world priors can be effectively transferred to trajectory generation for action generalization. 
In contrast, our work takes a planning-oriented perspective: \textsc{UNIVERSE} uses future video prediction as dense training-time supervision for trajectory generation, rather than treating video generation only as a standalone simulation task.

\subsection{World--Action Models for Autonomous Driving Control}

Beyond pure scene generation, world models have increasingly been incorporated into autonomous-driving planning and control. 
Representative methods include Driving into the Future~\cite{wang2024driving}, DrivingGPT~\cite{chen2025drivinggpt}, latent-world-model-assisted driving~\cite{li2024enhancing}, Doe-1~\cite{zheng2024doe}, ReSim~\cite{yang2025resim}, WoTE~\cite{li2025end_wote}, DriveLaW~\cite{xia2025drivelaw}, Epona~\cite{zhang2025epona}, Policy World Model~\cite{zhao2025forecasting}, and DriveVLA-W0~\cite{li2025drivevlaw0}. 
Beyond these driving models, video-based robotic planners and policies also exploit visual dynamics for transferable control~\cite{pai2025mimic,liang2025video,kim2026cosmos}. 
Meanwhile, recent embodied WAMs and video-action models, such as DreamZero~\cite{ye2026world}, GigaWorld-Policy~\cite{ye2026gigaworldpolicy}, LingBot-VA~\cite{lingbot-va2026}, Unified Video Action Model~\cite{li2025unified}, Unified World Models~\cite{zhu2025unified}, DreamGen~\cite{jang2025dreamgen}, VideoVLA~\cite{shen2025videovla}, and dual-stream diffusion~\cite{won2025dual}, further suggest that modeling future observations can improve robot control and generalization beyond the training domain.
Existing WAMs differ mainly in how future video prediction is coupled with control.
Dual-stream diffusion~\cite{won2025dual} maintains modality-specific pathways linked by shared cross-modal attention. 
Imagine-then-plan methods retain video rollout in the online loop, offering interpretable future imagination but incurring diffusion or autoregressive latency. 
DriveVA~\cite{liu2026driveva} jointly denoises future videos and ego trajectories in a shared DiT with direct future-modality interaction.
Fast-WAM~\cite{yuan2026fast} uses a pretrained video DiT and a separately initialized action-expert DiT, while Unified Video Action Model~\cite{li2025unified} uses a joint video-action latent representation with separate lightweight diffusion heads.
\textsc{UNIVERSE} builds on DriveVA's shared-parameter foundation but redesigns its future-token dependencies with a visibility mask, enabling independent output modes without requiring future-video tokens for trajectory inference.

\section{Method}
\label{sec:method}

\begin{figure}[t]
    \centering
    \includegraphics[width=1.0\linewidth]{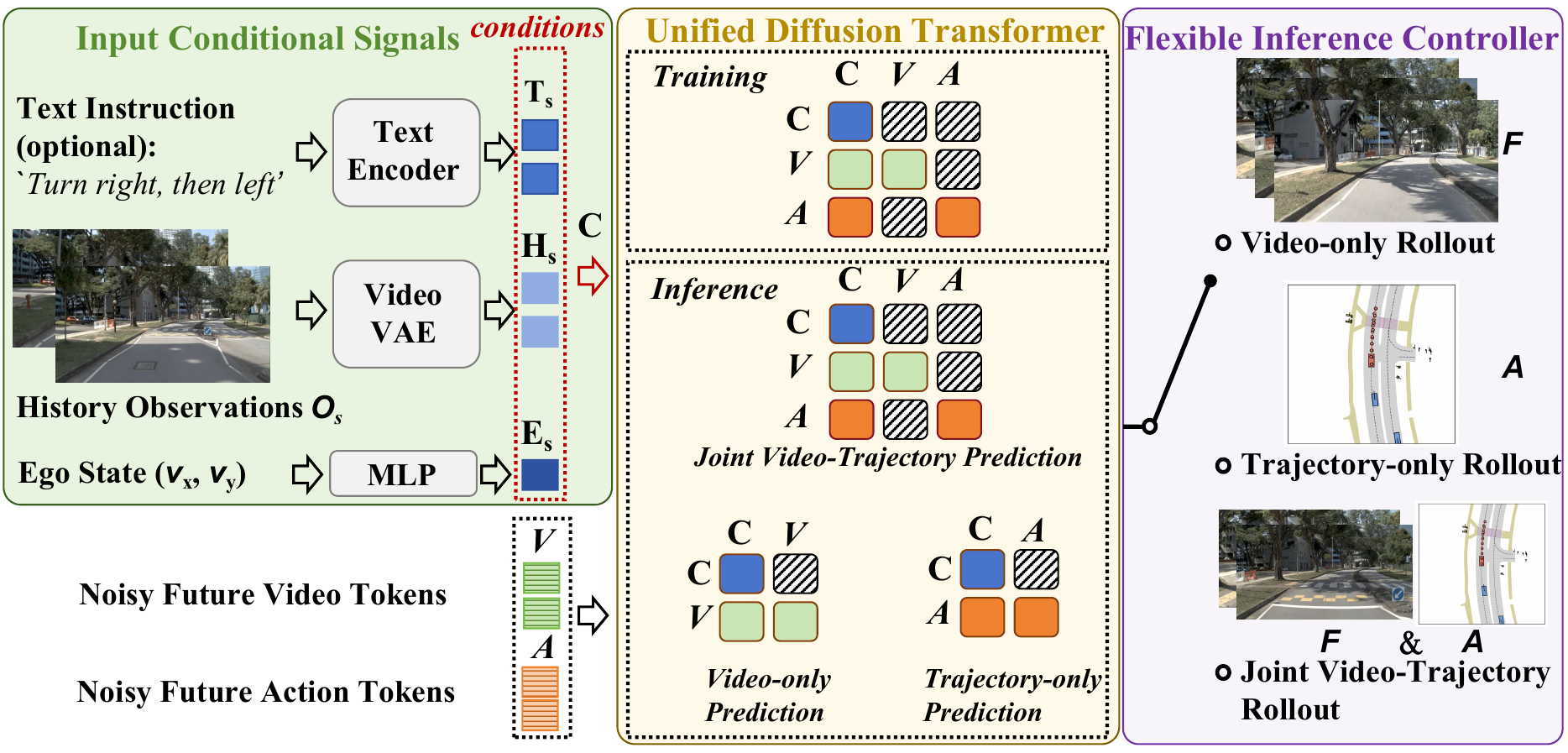}
    \caption{\textbf{Overall pipeline of \ourmethod.}
    Given history observations, ego velocity $(v_x,v_y)$, and language instructions, \ourmethod encodes conditional tokens with a text encoder and video VAE~\cite{wan2025}. 
    A unified DiT~\cite{peebles2023DIT} jointly trains future video latents and action tokens, while the Modality-Decoupling Visibility Mask blocks future video-action attention but preserves shared historical context. 
    The model supports joint, video-only, and trajectory-only prediction.}
    \label{fig:pipeline}
    \vspace{-0.3cm}
\end{figure}

\subsection{Unified Diffusion Transformer}
\label{sec:unified}

\textbf{Problem Definition.}
As shown in Fig.~\ref{fig:pipeline}, given $l$-frame history observations 
$\mathcal{O}_s=\{\mathbf{F}_{s-l+1},\ldots,\mathbf{F}_{s}\}$, current ego velocity $(v_x,v_y)$, and optional text instruction $\mathcal{T}$, \ourmethod predicts future actions and future visual rollouts either jointly or as a single modality.
The outputs are defined as
\begin{equation}
    \mathcal{A}_{s+1:s+K}
    =
    \{\boldsymbol{a}_{s+i}\in\mathbb{R}^3\}_{i=1}^{K},
    \qquad
    \mathcal{F}_{s+1:s+N}
    =
    \{\mathbf{F}_{s+j}\}_{j=1}^{N},
    \label{eq:problem_output}
\end{equation}
where each action $\boldsymbol{a}_{s+i}$ encodes the ego-vehicle $(x,y)$ position and yaw angle, and $\mathcal{F}_{s+1:s+N}$ describes the anticipated visual evolution of the same driving segment.
In practice, we predict latent visual and action representations following~\cite{zhang2025epona}, without using history trajectories as input.

\textbf{Architecture.}
A frozen text encoder from Wan2.2-TI2V-5B~\cite{wan2025} encodes the instruction $\mathcal{T}$ into latent tokens $\mathbf{T}_{s}\in\mathbb{R}^{L_T\times d}$.
A 3D-causal VAE adapted from Wan2.2-TI2V-5B encodes history observations into visual latents $\mathcal{H}^{\mathrm{his}}_s=\{\mathbf{H}_{s-l+1},\ldots,\mathbf{H}_{s}\}$, where $\mathbf{H}\in\mathbb{R}^{L_V\times d}$.
The ego velocity is embedded by an MLP, and the self-attention context tokens are constructed as
\begin{equation}
    \mathbf{T}_s=\mathrm{Enc}_{\mathrm{text}}(\mathcal{T}), 
    \qquad
    \mathcal{H}^{\mathrm{his}}_s=\mathrm{VAE}(\mathcal{O}_s),
    \qquad
    \mathbf{E}_s=\mathrm{MLP}(v_x,v_y),
    \qquad
    \mathcal{C}_s=\mathcal{H}^{\mathrm{his}}_s\cup \mathbf{E}_s .
    \label{eq:context_tokens}
\end{equation}
Text tokens $\mathbf{T}_{s}$ are injected through cross-attention.

For generation, future video latent tokens $\mathcal{V}_{s+1:s+N}$ and future trajectory tokens $\mathcal{A}_{s+1:s+K}$ are projected into the same hidden dimension $d$ with modality-specific input projections, modality embeddings, and temporal positional embeddings.
The DiT blocks, including self-attention and feed-forward layers, are shared by both modalities, while only the input projections and final prediction heads are modality-specific.
The unified prediction process is written as
\begin{equation}
    (\widehat{\mathbf{v}}^{\mathrm{vid}}, \widehat{\mathbf{v}}^{\mathrm{act}})
    =
    \bigl(
    h_{\mathrm{vid}}(\mathbf{Z}^{\mathrm{vid}}),
    h_{\mathrm{act}}(\mathbf{Z}^{\mathrm{act}})
    \bigr),
    \quad
    [\mathbf{Z}^{\mathrm{vid}},\mathbf{Z}^{\mathrm{act}}]
    =
    \mathrm{DiT}_{\theta}
    \bigl(
    \mathcal{V}_{s+1:s+N},
    \mathcal{A}_{s+1:s+K},
    \mathcal{C}_s,
    \mathbf{T}_s
    \bigr),
    \label{eq:unified_dit}
\end{equation}
where $h_{\mathrm{vid}}$ and $h_{\mathrm{act}}$ are modality-specific prediction heads for future video latents and trajectory tokens, respectively.
Thus, future-video prediction updates the same generative backbone used for trajectory generation instead of regularizing a separated planner.

\subsection{Modality-Decoupling Visibility Mask}
\label{sec:mask}

Generating future videos and trajectories in one transformer can create \textit{future-modality dependence}: trajectory tokens may rely on future visual tokens absent during trajectory-only inference, while video tokens may depend on future trajectory tokens. The issue is a training--inference dependency mismatch, rather than access to clean future ground truth at test time.
We therefore introduce a binary self-attention mask $M\in\{0,1\}^{L\times L}$ over context tokens $\mathcal{C}_s$, future video tokens $\mathcal{V}$, and future trajectory tokens $\mathcal{A}$, where $M(i,j)=1$ allows token $i$ to attend to token $j$ and $M(i,j)=0$ blocks this attention.
Text tokens are excluded since they are injected through cross-attention.

The mask allows future video and trajectory tokens to attend to historical context, i.e., $M(i,j)=1$ for $i\in\mathcal{V}\cup\mathcal{A}$ and $j\in\mathcal{C}_s$, but prevents context tokens from attending back to future targets.
It also blocks future video-action attention in both directions, i.e., $M(i,j)=0$ for $i\in\mathcal{V},j\in\mathcal{A}$ or $i\in\mathcal{A},j\in\mathcal{V}$, while preserving full attention within $\mathcal{C}_s$, $\mathcal{V}$, and $\mathcal{A}$.
This design shares historical context across modalities, blocks cross-modality future-token dependencies, and maintains intra-modality coherence for video generation and trajectory denoising.

\subsection{Flexible Inference Controller}
\label{sec:controller}

The proposed mask enables flexible inference by instantiating only the requested future token groups. 
For trajectory-only rollout, future video tokens are removed and the model denoises $\mathcal{A}_{s+1:s+K}$ conditioned on $\mathcal{C}_s$ and $\mathbf{T}_s$. 
For video-only rollout, future trajectory tokens are removed and the model denoises $\mathcal{V}_{s+1:s+N}$ from the same conditions. 
Because trajectory tokens are already blocked from future video tokens during training, removing the video branch at inference does not change their attention dependency, allowing trajectory-only inference to bypass future-video diffusion and improve efficiency by $4.3\times$.

For joint video-action rollout, both future token groups are instantiated and denoised within the same shared DiT under the Modality-Decoupling Visibility Mask. 
Although future video and trajectory tokens remain mutually blocked, they are aligned through the shared denoising backbone, common historical context, and joint co-training, enabling interpretable rollout while preserving the attention dependencies used for trajectory-only deployment.

\subsection{Training Objective}
\label{sec:objective}

We train \ourmethod with flow matching on both future trajectory tokens and future video latents. 
For a clean target sample $\mathbf{Y}_0^m$ from modality $m\in\{\mathrm{act},\mathrm{vid}\}$, we sample a flow time $p\sim\mathcal{U}(0,1)$ and Gaussian noise $\boldsymbol{\epsilon}^m\sim\mathcal{N}(0,I)$. 
The noisy sample and target velocity are
\begin{equation}
    \mathbf{Y}_p^m=(1-p)\boldsymbol{\epsilon}^m+p\mathbf{Y}_0^m,
    \qquad
    \dot{\mathbf{Y}}_p^m=\mathbf{Y}_0^m-\boldsymbol{\epsilon}^m .
\end{equation}

The noisy future trajectory and video tokens are jointly fed into the shared DiT with the visibility mask:
\begin{equation}
    (\hat{\mathbf{v}}_{\theta}^{\mathrm{act}}, 
    \hat{\mathbf{v}}_{\theta}^{\mathrm{vid}})
    =
    f_{\theta}
    (
    \mathbf{Y}_p^{\mathrm{act}},
    \mathbf{Y}_p^{\mathrm{vid}},
    p,
    \mathcal{C}_s,
    \mathbf{T}_s;
    M
    ),
\end{equation}
where $f_{\theta}$ denotes the shared DiT with modality-specific prediction heads. 
The final objective is
\begin{equation}
    \mathcal{L}
    =
    \mathcal{L}_{\mathrm{act}}
    +
    \lambda_{\mathrm{vid}}\mathcal{L}_{\mathrm{vid}},
    \qquad
    \mathcal{L}_{m}
    =
    \mathbb{E}
    \left[
    \left\|
    \hat{\mathbf{v}}_{\theta}^{m}
    -
    \dot{\mathbf{Y}}_p^{m}
    \right\|_2^2
    \right],
    \quad
    m\in\{\mathrm{act},\mathrm{vid}\}.
\end{equation}
Each modality loss is independently averaged over its batch and token dimensions before weighting, so the number of video tokens does not directly determine its contribution.
We use $\lambda_{\mathrm{vid}}=1.0$ by default; Sec.~\ref{sec:pretraining_sensitivity_main} reports sensitivity to $0.5$ and $2.0$.
Both objectives update the same DiT parameters.

\subsection{\ourmethod Variants}
\label{sec:variants}

We instantiate several variants to isolate the effects of video co-training, mask-based modality decoupling, and shared DiT parameterization. 
All variants use the same observations, trajectory targets, tokenization, training schedule, and evaluation protocol unless otherwise specified.

\textbf{\ourmethod.}
The full model co-trains future video latents and trajectory tokens within one shared DiT under the Modality-Decoupling Visibility Mask.

\textbf{\ourmethod w.o. Mask.}
To instantiate a controlled DriveVA-style coupled dependency structure~\citep{liu2026driveva}, we remove the Modality-Decoupling Visibility Mask, allowing future video and trajectory tokens to attend to each other.
This keeps the unified DiT and joint objective, but requires coupled video-action generation at inference.

\textbf{\ourmethod w.o. video co-train.}
This variant keeps the architecture, tokenization, mask, and inference procedure unchanged, but removes video supervision by setting $\lambda_{\mathrm{vid}}=0$, namely $\mathcal{L}=\mathcal{L}_{\mathrm{act}}$.

\textbf{\ourmethod Two-DiT.}
Following Fast-WAM~\citep{yuan2026fast}, we replace the shared DiT with two separated DiTs for video and trajectory denoising.
They share the same conditions and losses but not transformer parameters, preserving video co-training while removing shared video-action parameterization.

\section{Experiments}
\label{sec:experiments}

\subsection{Datasets and Implementation Details}
\label{sec:exp_setup}

\noindent\textbf{NAVSIM v1.}
Following prior works~\citep{zhang2025epona,xia2025drivelaw}, we adopt NAVSIM\,v1~\citep{dauner2024navsim}, built upon OpenScene~\citep{openscene2023}, as our main non-reactive simulation-based planning benchmark.
It reports No Collision (NC), Drivable Area Compliance (DAC), Time-to-Collision (TTC), Comfort (C.), and Ego Progress (EP), aggregated as
$\mathrm{PDMS}=\mathrm{NC}\times \mathrm{DAC}\times (5\mathrm{EP}+5\mathrm{TTC}+2\mathrm{C.})/12$.
Unless otherwise specified, our model only uses front-view camera input without map, box, depth, occupancy, or LiDAR supervision.

\noindent\textbf{nuScenes.}
For cross-dataset evaluation, we directly test the NAVSIM-trained model on the nuScenes validation split of 150 scenes from the official 1,000-scene dataset~\citep{caesar2020nuscenes}.
Following prior planning works~\citep{hu2023planning,jiang2023vad}, we report L2 displacement error and collision rate at 1\,s, 2\,s, and 3\,s horizons.
We also report FID and FVD for future-video generation quality following driving world-model literature~\citep{wang2024drivedreamer,gao2024vista}.

\noindent\textbf{Bench2Drive.}
Bench2Drive~\citep{jia2024bench2drive} is a CARLA\,v2 benchmark~\citep{dosovitskiy2017carla} with diverse interactive scenarios and routes.
We use it for real-to-simulation zero-shot transfer by directly evaluating the NAVSIM-trained model on its validation split, where the model must handle shifts in visual appearance, vehicle dynamics, and agent behaviors~\citep{hu2023simulation}.

\noindent\textbf{Implementation Details.}
Training is conducted on NVIDIA H20 GPUs with AdamW, a learning rate of $10^{-4}$, weight decay of $0.01$, and distributed bf16 mixed-precision training.
We first train with a batch size of $80$ for $20$k steps for fast convergence, and then continue training for another $10$k steps with an effective batch size of $640$ using gradient accumulation.
We use a $1$k-step linear warm-up from $10^{-3}$ of the base learning rate, followed by a constant learning-rate schedule.
The training objective combines a flow-matching loss for future video supervision and a trajectory prediction loss.
During inference, we use the trajectory-only rollout mode with $2$ flow sampling steps unless otherwise stated.
Wan2.2-TI2V-5B~\citep{wan2025} is used as the pre-trained video generation backbone.
Each training sample contains $l=4$ history frames and $N=8$ future frames at a resolution of $832\times480$.
The DiT attention/FFN parameters, modality-specific input projections, and prediction heads are trainable; the text encoder and VAE are frozen.
All zero-shot results use one NAVSIM-trained checkpoint without target-domain fine-tuning or adaptation. The Bench2Drive L2 and collision results are open-loop; input settings and preprocessing are detailed in Sec.~\ref{sec:supp_protocol}.

\begin{table}[t!]
    \centering
    \caption{\textbf{Performance comparison on NAVSIM \textit{Navtest} using non-reactive simulation-based metrics.}
    Methods are grouped by whether they employ an explicit world model: \textit{Traditional End-to-End Methods} and \textit{World Model Methods}.}
    \setlength{\tabcolsep}{6pt}
    \label{tab:navsim1}
    \resizebox{0.98\textwidth}{!}{%
    \begin{tabular}{@{}l|c|cc|cc|cccc@{}}
        \toprule
        \textbf{Method} & \textbf{Ref} & \textbf{Image} & \textbf{Lidar} &
        \textbf{NC$\uparrow$} & \textbf{DAC$\uparrow$} &
        \textbf{TTC$\uparrow$} & \textbf{Comf.$\uparrow$} & \textbf{EP$\uparrow$} & \textbf{PDMS$\uparrow$} \\
        \midrule

        \rowcolor{gray!30}\multicolumn{10}{@{}l}{\raggedright \textit{Traditional End-to-End Methods}} \\
        VADv2-$\mathcal{V}_{\text{8192}}$~\citep{ICLR2026_70925fe2} & ICLR'26 & \cmark &  & 97.2 & 89.1 & 91.6 & \textbf{100} & 76.0 & \cellcolor{gray!30}80.9 \\
        UniAD~\citep{hu2023planning} & CVPR'23 & \cmark &  & 97.8 & 91.9 & 92.9 & \textbf{100} & 78.8 & \cellcolor{gray!30}83.4 \\
        TransFuser~\citep{chitta2022transfuser} & TPAMI'23 & \cmark & \cmark & 97.7 & 92.8 & 92.8 & \textbf{100} & 79.2 & \cellcolor{gray!30}84.0 \\
        PARA-Drive~\citep{weng2024drive} & CVPR'24 & \cmark &  & 97.9 & 92.4 & 93.0 & 99.8 & 79.3 & \cellcolor{gray!30}84.0 \\
        ReCogDrive-IL~\citep{li2025recogdrive} & ICLR'26 & \cmark &  & 98.1 & 94.7 & 94.2 & \textbf{100} & 80.9 & \cellcolor{gray!30}86.5 \\
        DiffusionDrive~\citep{liao2025diffusiondrive} & CVPR'25 & \cmark & \cmark & 98.2 & 96.2 & 94.7 & \textbf{100} & 82.2 & \cellcolor{gray!30}88.1 \\
        \midrule

        \rowcolor{gray!30} \multicolumn{10}{@{}l}{\raggedright \textit{World Model Methods}} \\
        DrivingGPT~\citep{chen2025drivinggpt} & ICCV'25 & \cmark &  & 98.9 & 90.7 & 94.9 & 95.6 & 79.7 & \cellcolor{gray!30}82.4 \\
        LAW~\citep{li2024enhancing} & ICLR'25 & \cmark &  & 96.4 & 95.4 & 88.7 & 99.9 & 81.7 & \cellcolor{gray!30}84.6 \\
        Epona~\citep{zhang2025epona} & ICCV'25 & \cmark &  & 97.9 & 95.1 & 93.8 & 99.9 & 80.4 & \cellcolor{gray!30}86.2 \\
        Resim~\citep{yang2025resim} & NeurIPS'25 & \cmark &  & -- & -- & -- & -- & -- & \cellcolor{gray!30}86.6 \\
        WoTE~\citep{li2025end_wote} & ICCV'25 & \cmark & \cmark & 98.5 & 96.8 & 94.9 & 99.9 & 81.9 & \cellcolor{gray!30}88.3 \\
        DriveVLA-W0~\citep{li2025drivevlaw0} & ICLR'26 & \cmark &  & 98.4 & 95.3 & 95.2 & \textbf{100} & 80.9 & \cellcolor{gray!30}87.2 \\
        PWM~\citep{zhao2025forecasting} & NeurIPS'25 & \cmark &  & 98.6 & 95.9 & 95.4 & \textbf{100} & 81.8 & \cellcolor{gray!30}88.1 \\
        DriveLaW~\citep{xia2025drivelaw} & CVPR'26 & \cmark &  & 99.0 & 97.1 & 96.7 & \textbf{100} & 81.3 & \cellcolor{gray!30}89.1 \\
        DriveVA~\citep{liu2026driveva} & ECCV'26 & \cmark &  & \textbf{99.2} & 97.5 & \textbf{98.7} & \textbf{100} & 83.5 & \cellcolor{gray!30}90.9 \\
        \midrule

        \rowcolor{gray!30} \multicolumn{10}{@{}l}{\raggedright \textit{UNIVERSE Variants}} \\
        \midrule
        UNIVERSE & - & \cmark &  & 99.1 & \textbf{97.6} & 98.5 & \textbf{100} & \textbf{83.6} & \cellcolor{gray!30}\textbf{91.0} \\
        UNIVERSE w.o. Mask & - & \cmark &  & \textbf{99.2} & 97.5 & \textbf{98.7} & \textbf{100} & 83.5 & \cellcolor{gray!30}90.9 \\
        UNIVERSE w.o. video co-train & - & \cmark &  & 98.8 & 95.5 & 98.2 & 99.7 & 80.4 & \cellcolor{gray!30}88.2 \\
        UNIVERSE Two-DiT & - & \cmark &  & 98.9 & 96.2 & 98.3 & 99.9 & 82.7 & \cellcolor{gray!30}89.6 \\
        \bottomrule
    \end{tabular}%
    }
    \vspace{-0.3cm}
\end{table}

\subsection{Comparison Results on NAVSIM v1}
\label{sec:navsim_result}

Table~\ref{tab:navsim1} reports non-reactive simulation-based planning results on NAVSIM v1.
UNIVERSE achieves a PDMS of $91.0$, outperforming prior traditional end-to-end planners and recent world-model-based planners.
Compared with DiffusionDrive~\citep{liao2025diffusiondrive}, our method improves PDMS from $88.1$ to $91.0$.
Compared with WoTE~\citep{li2025end_wote}, which uses both camera and LiDAR inputs, UNIVERSE still obtains a higher PDMS using only the front-view camera.
Compared with DriveLaW~\citep{xia2025drivelaw}, UNIVERSE improves PDMS from $89.1$ to $91.0$, with higher DAC, TTC, and EP scores under non-reactive simulation.
DriveVA~\citep{liu2026driveva} achieves $90.9$ PDMS, compared with $91.0$ for UNIVERSE.

These results support our central hypothesis: future-video supervision is more effective for planning when it directly updates the same generative parameters used for trajectory denoising.
Instead of using future videos only as an auxiliary output or a separate imagination module, UNIVERSE co-trains video latents and trajectory tokens in one shared DiT, allowing dense visual dynamics to regularize action generation.
The UNIVERSE variant rows are included to preview the controlled comparison results; since these variants are designed to isolate architectural factors rather than serve as standalone baselines, we defer their detailed discussion to Sec.~\ref{sec:ablation}.

\begin{figure}[t]
    \centering
    \includegraphics[width=1.0\linewidth]{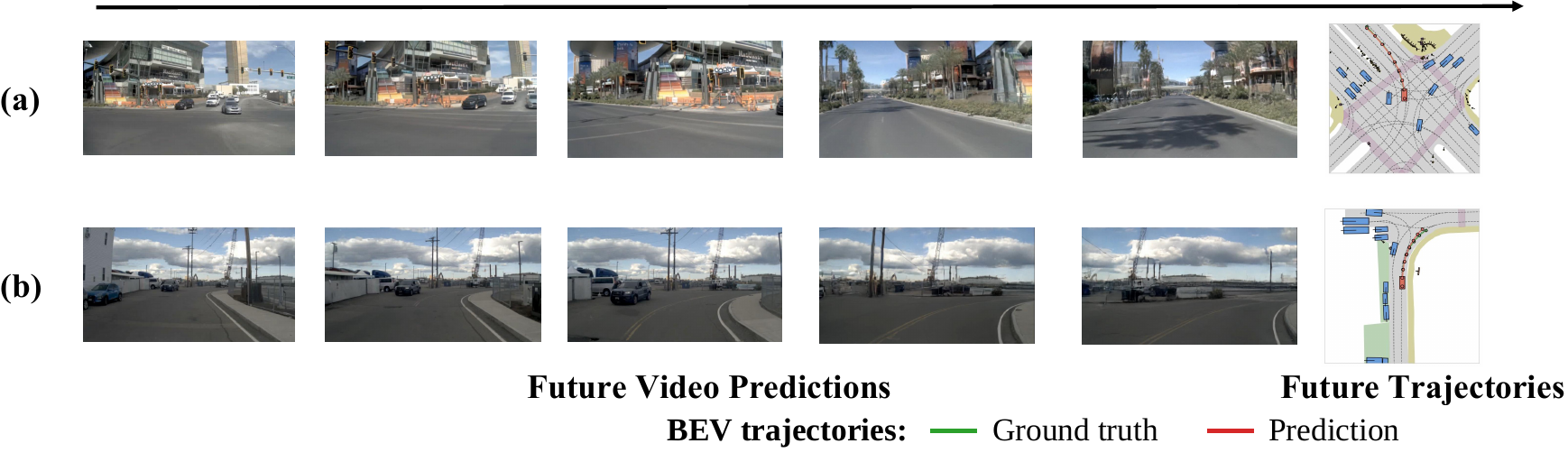}
    \caption{\textbf{Qualitative future video and trajectory generation on NAVSIM v1.}
    The rightmost BEV panels compare the ground-truth trajectory (green) and prediction (red); the video panels show predictions rather than paired ground-truth video rows.
    Paired ground-truth/generated video examples and DPVO~\citep{teed2023deep} analysis are provided in Secs.~\ref{sec:supp_dpvo} and~\ref{sec:supp_visual}.}
    \label{fig:visual1}
    \vspace{-0.15cm}
\end{figure}

\begin{figure}[t]
    \centering
    \includegraphics[width=1.0\linewidth]{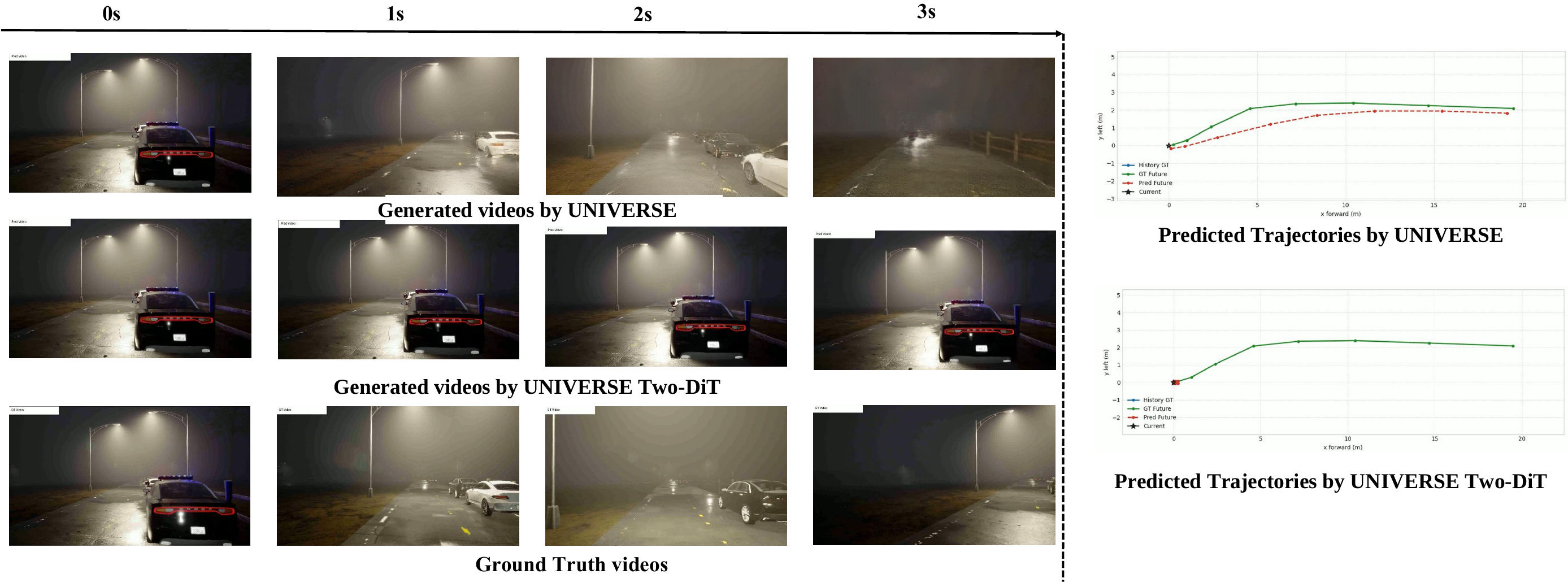}
    \caption{\textbf{Zero-shot driving-intention comparison.}
    UNIVERSE is evaluated on Bench2Drive without fine-tuning.
    Red and green curves denote \textcolor{red}{predicted} and \textcolor{green}{ground-truth} trajectories, respectively.
    When a stationary vehicle blocks the lane ahead, UNIVERSE predicts a bypass trajectory, while the Two-DiT variant tends to stop behind it. Enlarged trajectory plots appear in Fig.~\ref{fig:supp_trajectory_detail}.}
    \label{fig:visual_compare_two_dit}
    \vspace{-0.25cm}
\end{figure}

% v4: Only two inconsistent Avg. cells are recomputed from displayed 1s/2s/3s values.
% PWM/nuScenes: (0.12+0.15+0.86)/3 = 0.376666... -> 0.38 percent.
% w.o. Mask/Bench2Drive: (1.38+1.97+2.65)/3 = 2.00 percent.
% These are arithmetic recalculations, not new experiment or log-verified results.
\begin{table*}[t]
\setlength{\tabcolsep}{1.5pt}
\caption{\textbf{Zero-shot end-to-end motion planning performance on nuScenes~\citep{caesar2020nuscenes} and Bench2Drive (CARLA)~\citep{jia2024bench2drive}.}
The UNIVERSE variants use a NAVSIM-trained checkpoint and are directly tested on the target datasets without fine-tuning. Results for external baselines follow their respective cited protocols. Both target-dataset evaluations are open-loop. $\dagger$ indicates averages recomputed from the displayed 1s, 2s, and 3s collision rates.}
\centering
\resizebox{0.98\linewidth}{!}{
\begin{tabular}{l|cc|cccc|cccc|cccc|cccc}
\toprule
\multirow{2}{*}{Method} &
\multirow{2}{*}{Finetune} &
\multirow{2}{*}{Ref} &
\multicolumn{8}{c|}{nuScenes} &
\multicolumn{8}{c}{Bench2Drive (CARLA)} \\
\cline{4-19}
&&&
\multicolumn{4}{c|}{L2 (m) $\downarrow$} &
\multicolumn{4}{c|}{Collision (\%) $\downarrow$} &
\multicolumn{4}{c|}{L2 (m) $\downarrow$} &
\multicolumn{4}{c}{Collision (\%) $\downarrow$} \\
&&&
1s & 2s & 3s & \cellcolor{gray!30}Avg. &
1s & 2s & 3s & \cellcolor{gray!30}Avg. &
1s & 2s & 3s & \cellcolor{gray!30}Avg. &
1s & 2s & 3s & \cellcolor{gray!30}Avg. \\
\midrule
\rowcolor{gray!30}
\multicolumn{19}{@{}l}{\raggedright \textit{VLA-World Model Methods}} \\
DriveVLA-W0~\citep{li2025drivevlaw0}
& \xmark
& ICLR'26
& 0.43 & 1.26 & 2.60 & \cellcolor{gray!30}1.43
& 0.22 & 0.66 & 1.42 & \cellcolor{gray!30}0.77
& 1.01 & 2.77 & 5.22 & \cellcolor{gray!30}3.00
& 1.49 & 2.53 & 3.53 & \cellcolor{gray!30}2.52 \\
\midrule
\rowcolor{gray!30}
\multicolumn{19}{@{}l}{\raggedright \textit{World Model Methods}} \\
PWM~\citep{zhao2025forecasting}
& \xmark
& NeurIPS'25
& 2.06 & 3.91 & 6.00 & \cellcolor{gray!30}3.99
& 0.12 & 0.15 & 0.86 & \cellcolor{gray!30}$0.38^{\dagger}$
& 1.70 & 2.74 & 3.97 & \cellcolor{gray!30}2.80
& 4.01 & 3.73 & 3.53 & \cellcolor{gray!30}3.76 \\
\midrule
\rowcolor{gray!30}
% \multicolumn{19}{@{}l}{\raggedright \textit{Published Video--Action Predecessor}} \\
% DriveVA~\citep{liu2026driveva}
% & \xmark
% & ECCV'26
% & 0.33 & 0.76 & 1.43 & \cellcolor{gray!30}0.84
% & 0.00 & 0.07 & 0.12 & \cellcolor{gray!30}0.06
% & 0.69 & 1.29 & 2.03 & \cellcolor{gray!30}1.33
% & 1.38 & 1.97 & 2.65 & \cellcolor{gray!30}1.79 \\
% \midrule
% \rowcolor{gray!30}
\multicolumn{19}{@{}l}{\raggedright \textit{UNIVERSE Variants}} \\
\midrule
UNIVERSE
& \xmark
& --
& \textbf{0.31} & \textbf{0.76} & 1.44 & \cellcolor{gray!30}\textbf{0.83}
& \textbf{0.00} & \textbf{0.03} & \textbf{0.05} & \cellcolor{gray!30}\textbf{0.03}
& \textbf{0.67} & \textbf{1.29} & \textbf{2.03} & \cellcolor{gray!30}\textbf{1.33}
& \textbf{0.75} & \textbf{1.12} & \textbf{1.70} & \cellcolor{gray!30}\textbf{1.19} \\
UNIVERSE w.o. Mask
& \xmark
& --
& 0.33 & \textbf{0.76} & \textbf{1.43} & \cellcolor{gray!30}0.84
& \textbf{0.00} & 0.07 & 0.12 & \cellcolor{gray!30}0.06
& 0.69 & \textbf{1.29} & \textbf{2.03} & \cellcolor{gray!30}1.34
& 1.38 & 1.97 & 2.65 & \cellcolor{gray!30}$2.00^{\dagger}$ \\
UNIVERSE w.o. video co-train
& \xmark
& --
& 0.48 & 0.94 & 1.65 & \cellcolor{gray!30}1.02
& 0.37 & 0.31 & 0.38 & \cellcolor{gray!30}0.35
& 0.75 & 1.38 & 2.15 & \cellcolor{gray!30}1.43
& 1.07 & 1.47 & 2.16 & \cellcolor{gray!30}1.57 \\
UNIVERSE Two-DiT
& \xmark
& --
& 0.44 & 0.81 & 1.52 & \cellcolor{gray!30}0.92
& 0.17 & 0.22 & 0.25 & \cellcolor{gray!30}0.21
& 0.80 & 1.39 & 2.11 & \cellcolor{gray!30}1.43
& 1.30 & 1.83 & 2.32 & \cellcolor{gray!30}1.82 \\
\bottomrule
\end{tabular}}
\label{tab:zero_shot}
\vspace{-0.2cm}
\end{table*}

\subsection{Zero-Shot Generalization on nuScenes and Bench2Drive}
\label{sec:zero_shot}

\noindent\textbf{From NAVSIM to nuScenes.}
To evaluate cross-dataset action generalization, we directly test the NAVSIM-trained UNIVERSE on nuScenes without any fine-tuning.
As shown in Table~\ref{tab:zero_shot}, UNIVERSE achieves an average L2 error of $0.83$\,m and an average collision rate of $0.03\%$.
Compared with PWM~\citep{zhao2025forecasting}, which is also trained on NAVSIM and evaluated zero-shot, UNIVERSE reduces the average L2 error from $3.99$\,m to $0.83$\,m and the average collision rate from $0.38\%$ to $0.03\%$.
Compared with DriveVLA-W0~\citep{li2025drivevlaw0}, UNIVERSE also improves the average L2 error from $1.43$\,m to $0.83$\,m and the average collision rate from $0.77\%$ to $0.03\%$.
These results support improved transfer beyond NAVSIM; a source-domain fitting analysis and its interpretive limits are given in Sec.~\ref{sec:supp_generalization}.

\noindent\textbf{From NAVSIM to Bench2Drive.}
We further evaluate zero-shot transfer from real-world logs to simulation on Bench2Drive.
As reported in Table~\ref{tab:zero_shot}, UNIVERSE obtains an average L2 error of $1.33$\,m and an average collision rate of $1.19\%$.
Compared with PWM~\citep{zhao2025forecasting}, UNIVERSE reduces the average L2 error from $2.80$\,m to $1.33$\,m and the average collision rate from $3.76\%$ to $1.19\%$.
Fig.~\ref{fig:visual_compare_two_dit} further provides a qualitative zero-shot example: when a stationary vehicle blocks the lane ahead, UNIVERSE predicts a reasonable bypass maneuver, while the Two-DiT variant tends to stop behind the obstacle.
This indicates stronger driving-intention understanding under real-to-simulation domain shift.

\begin{table}[t]
\centering
\caption{\textbf{Comparison with closely related video-action methods.}
Fast-WAM is adapted from official code and trained on NAVSIM.}
\label{tab:closest_methods}
\small
\setlength{\tabcolsep}{4pt}
\begin{tabular}{@{}lccccc@{}}
\toprule
\multirow{2}{*}{Method} & NAVSIM & \multicolumn{2}{c}{nuScenes} & \multicolumn{2}{c}{Bench2Drive} \\
\cmidrule(lr){3-4}\cmidrule(l){5-6}
& PDMS$\uparrow$ & L2 (m)$\downarrow$ & Col. (\%)$\downarrow$ & L2 (m)$\downarrow$ & Col. (\%)$\downarrow$ \\
\midrule
Fast-WAM~\citep{yuan2026fast} & 89.8 & 0.89 & 0.23 & 1.45 & 1.85 \\
DriveVA~\citep{liu2026driveva} & 90.9 & 0.84 & 0.06 & \textbf{1.33} & 1.79 \\
\ourmethod & \textbf{91.0} & \textbf{0.83} & \textbf{0.03} & \textbf{1.33} & \textbf{1.19} \\
\bottomrule
\end{tabular}
\end{table}

\noindent\textbf{Comparison with close video-action models.}
Table~\ref{tab:closest_methods} complements the controlled variants with an adapted Fast-WAM baseline and DriveVA results.
\ourmethod preserves comparable planning accuracy while supporting trajectory-only inference with the same weights.

\begin{table*}[t]
\centering
\caption{\textbf{Quantitative evaluation of future-video generation on the nuScenes validation set~\citep{caesar2020nuscenes}.}
UNIVERSE achieves competitive visual generation quality while being optimized for planning-oriented video-action co-training.}
\label{tab:nuscenes_video_gen}
\small
\setlength{\tabcolsep}{3.2pt}
\renewcommand{\arraystretch}{1.05}
\resizebox{\textwidth}{!}{
\begin{tabular}{lccccccc}
\toprule
\textbf{Metric}
& DriveGAN~\citep{kim2021drivegan}
& DriveDreamer~\citep{wang2024drivedreamer}
& DrivingGPT~\citep{chen2025drivinggpt}
& DrivingWorld~\citep{hu2024drivingworld}
& Vista~\citep{gao2024vista}
& Epona~\citep{zhang2025epona}
& \textbf{Ours} \\
\midrule
FID $\downarrow$
& 73.4
& 52.6
& 12.8
& 7.4
& 6.9
& 7.5
& \textbf{6.7} \\
FVD $\downarrow$
& 502.3
& 452.0
& 142.6
& 90.9
& 89.4
& 82.8
& \textbf{81.7} \\
\bottomrule
\end{tabular}}
\vspace{-0.3em}
\end{table*}

\subsection{Future-Video Generation Quality}
\label{sec:video_generation}

Although motion planning is our primary goal, UNIVERSE also generates high-quality future videos.
As shown in Table~\ref{tab:nuscenes_video_gen}, it achieves the best FID of $6.7$ and FVD of $81.7$ on nuScenes among the compared driving video-generation methods.
This indicates that the video branch learns realistic and temporally coherent scene evolution, while the shared DiT allows such video dynamics modeling to regularize trajectory generation.

\begin{table}[t]
\centering
\small
\setlength{\tabcolsep}{8pt}
\renewcommand{\arraystretch}{1.05}

\begin{minipage}[t]{\linewidth}
\centering
\captionof{table}{\textbf{Video-action consistency.}
DPVO~\citep{teed2023deep} 4s ego-motion L2 (m); protocol in Sec.~\ref{sec:supp_dpvo}.}
\label{tab:dpvo_consistency}
\begin{tabular}{l|ccc}
\toprule
\textbf{Method}
& \textbf{NAVSIM$\downarrow$}
& \textbf{nuScenes$\downarrow$}
& \cellcolor{gray!30}\textbf{Avg.$\downarrow$} \\
\midrule
GT reference
& 0.09
& 0.07
& \cellcolor{gray!30}0.08 \\
\midrule
UNIVERSE
& \textbf{0.12}
& \textbf{0.14}
& \cellcolor{gray!30}\textbf{0.13} \\
UNIVERSE w.o. Mask
& 0.16
& \textbf{0.14}
& \cellcolor{gray!30}0.15 \\
UNIVERSE Two-DiT
& 0.17
& 0.21
& \cellcolor{gray!30}0.19 \\
\bottomrule
\end{tabular}
\end{minipage}
\par\medskip
\begin{minipage}[t]{\linewidth}
\centering
\captionof{table}{\textbf{Computational efficiency.}
Trajectory inference latency, inference memory, and training memory.}
\label{tab:efficiency}
\begin{tabular}{l|ccc}
\toprule
\textbf{Method}
& \textbf{\shortstack{Latency\\(ms)$\downarrow$}}
& \textbf{\shortstack{Infer. Peak\\Mem. (GB)$\downarrow$}}
& \textbf{\shortstack{Train Peak\\Mem. (GB)$\downarrow$}} \\
\midrule
UNIVERSE
& \textbf{376}
& \textbf{23.3}
& \textbf{52.1} \\
UNIVERSE w.o. Mask
& 1623
& 26.3
& 54.4 \\
UNIVERSE Two-DiT
& 551
& 32.8
& 88.3 \\
\bottomrule
\end{tabular}
\end{minipage}
\end{table}

\subsection{Controlled Comparison with UNIVERSE Variants}
\label{sec:ablation}

The UNIVERSE variants are designed to answer which architectural factors are responsible for the observed action generalization.
All variants use the same observations, trajectory targets, tokenization, training schedule, and evaluation protocol, but differ in one key factor: whether future-video supervision is used, whether future video and trajectory tokens are decoupled by the visibility mask, and whether video and trajectory denoising share the same DiT parameters.
We compare them jointly across in-domain planning, zero-shot transfer, video-action consistency, inference latency, and memory consumption.

\noindent\textbf{Dense video supervision improves transfer.}
Removing future-video supervision reduces NAVSIM PDMS from $91.0$ to $88.2$ in Table~\ref{tab:navsim1}.
The degradation becomes more evident under domain shift: on nuScenes, the average L2 error increases from $0.83$\,m to $1.02$\,m and the collision rate increases from $0.03\%$ to $0.35\%$; on Bench2Drive, the collision rate increases from $1.19\%$ to $1.57\%$.
This confirms that future-video prediction is not merely an auxiliary visualization task, but provides dense dynamics supervision that improves transferable action generation.

\noindent\textbf{Shared denoising backbone matters.}
UNIVERSE Two-DiT preserves both video and trajectory objectives but separates them into two diffusion transformers.
Although it still benefits from video co-training at the system level, its NAVSIM PDMS drops to $89.6$, and its zero-shot performance is consistently weaker than the shared-DiT UNIVERSE.
On nuScenes, the average L2 error increases from $0.83$\,m to $0.92$\,m and the collision rate from $0.03\%$ to $0.21\%$; on Bench2Drive, the collision rate increases from $1.19\%$ to $1.82\%$.
Qualitatively, Fig.~\ref{fig:visual_compare_two_dit} shows that the Two-DiT variant tends to stop behind a stationary obstacle, while UNIVERSE predicts a bypass trajectory, indicating stronger driving-intention understanding from shared video-action denoising.
It also produces the largest video-action consistency error in Table~\ref{tab:dpvo_consistency} and the highest training peak memory in Table~\ref{tab:efficiency}.
These results show that video co-training is most effective when video and trajectory losses update the same generative backbone.

\noindent\textbf{Visibility decoupling enables flexible deployment.}
The no-mask variant obtains comparable in-domain performance with a PDMS of $90.9$, but this does not translate into better deployment behavior.
Because future video and trajectory tokens can attend to each other, the model relies on coupled video-action rollout and cannot remove one future modality without changing the learned dependency structure.
This leads to weaker zero-shot safety: the collision rate increases from $0.03\%$ to $0.06\%$ on nuScenes and from $1.19\%$ to $2.00\%$ on Bench2Drive.
It also increases trajectory inference latency from $376$\,ms to $1623$\,ms, since future-video denoising must be retained during rollout.
The visibility mask therefore preserves training--inference dependency consistency while enabling flexible trajectory-only deployment. Small submetric trade-offs are discussed in Sec.~\ref{sec:supp_tradeoffs}.

\noindent\textbf{Video-action consistency and efficiency.}
The qualitative results in Fig.~\ref{fig:visual1} show that generated future videos and predicted trajectories evolve consistently.
To quantify this alignment, we use DPVO~\citep{teed2023deep} to reconstruct ego-motion from ground-truth and generated future videos, align the reconstructed trajectories to their references with a similarity transform, and report the average L2 error over the future $4$\,s horizon.
As shown in Table~\ref{tab:dpvo_consistency}, UNIVERSE achieves an average error of $0.13$, close to the GT-reference error of $0.08$, while the no-mask and Two-DiT variants increase the error to $0.15$ and $0.19$.
Meanwhile, Table~\ref{tab:efficiency} shows that UNIVERSE achieves the lowest latency, inference memory, and training memory among the variants.
Together, these results support the coupling-by-parameters and decoupling-by-visibility design: video and trajectory objectives should share the same DiT parameters, while their future tokens should remain mutually invisible.

Additional comparisons of video backbones and NAVSIM Train--Navtest gaps are provided in Sec.~\ref{sec:supp_additional_experiments}, and horizon and sampling ablations are in Sec.~\ref{sec:horizon_steps}.

\subsection{Pretraining and Sensitivity Analyses}
\label{sec:pretraining_sensitivity_main}

\noindent\textbf{Pretrained initialization.}
Table~\ref{tab:supp_pretraining} compares the same 5B architecture trained from scratch with one initialized from pretrained video-generation weights.
The gain from $63.5$ to $91.0$ NAVSIM PDMS shows that pretrained initialization substantially contributes to performance, alongside the architectural effects studied above.

\begin{table}[htbp]
\centering
\caption{\textbf{Effect of pretrained initialization.} Both rows use the 5B UNIVERSE architecture. Target-dataset metrics are average open-loop L2 (m) and collision rate (\%).}
\label{tab:supp_pretraining}
\small
\setlength{\tabcolsep}{6pt}
\begin{tabular}{@{}lccccc@{}}
\toprule
Initialization & NAVSIM & \multicolumn{2}{c}{nuScenes} & \multicolumn{2}{c}{Bench2Drive} \\
\cmidrule(lr){2-2}\cmidrule(lr){3-4}\cmidrule(l){5-6}
 & PDMS$\uparrow$ & L2$\downarrow$ & Col.$\downarrow$ & L2$\downarrow$ & Col.$\downarrow$ \\
\midrule
From scratch & 63.5 & 1.32 & 1.23 & 2.48 & 4.85 \\
Pretrained weights & \textbf{91.0} & \textbf{0.83} & \textbf{0.03} & \textbf{1.33} & \textbf{1.19} \\
\bottomrule
\end{tabular}
\end{table}

\noindent\textbf{Video-loss weight sensitivity.}
Table~\ref{tab:main_loss_weight} studies the effect of the video-loss weight $\lambda_{\mathrm{vid}}$ while fixing the Wan2.2-TI2V-5B backbone.
The default setting $\lambda_{\mathrm{vid}}=1.0$ gives the best overall balance across source and cross-domain evaluations.

\begin{table}[htbp]
\centering
\caption{\textbf{Video-loss weight sensitivity.} Wan2.2-TI2V-5B is fixed; $\lambda_{\mathrm{vid}}=1.0$ is the default. L2 is in meters and collision is in percent.}
\label{tab:main_loss_weight}
\small
\setlength{\tabcolsep}{6pt}
\begin{tabular}{@{}lccccc@{}}
\toprule
$\lambda_{\mathrm{vid}}$ & NAVSIM & \multicolumn{2}{c}{nuScenes} & \multicolumn{2}{c}{Bench2Drive} \\
\cmidrule(lr){2-2}\cmidrule(lr){3-4}\cmidrule(l){5-6}
 & PDMS$\uparrow$ & L2$\downarrow$ & Col.$\downarrow$ & L2$\downarrow$ & Col.$\downarrow$ \\
\midrule
0.5 & 90.8 & 0.83 & 0.05 & 1.34 & \textbf{1.18} \\
1.0 (default) & \textbf{91.0} & \textbf{0.83} & \textbf{0.03} & \textbf{1.33} & 1.19 \\
2.0 & 90.9 & \textbf{0.83} & 0.05 & 1.36 & 1.21 \\
\bottomrule
\end{tabular}
\end{table}
\FloatBarrier

\section{Conclusion}
We present UNIVERSE, a video-action model that retains dense video supervision of shared DiT parameters while making future-video generation optional at deployment.
The Modality-Decoupling Visibility Mask separates future-token dependencies and enables trajectory-only, video-only, and joint inference with one set of weights.
Experiments show strong NAVSIM performance and zero-shot transfer to nuScenes and Bench2Drive, with $4.3\times$ faster trajectory inference than coupled rollout and lower memory use than the compared variants.
The results support shared-parameter training with dependency-consistent modality decoupling.

\clearpage
% AUTHOR CONFIRMATION REQUIRED: replace the visible placeholder below.
% No grant, employment, hardware support, or "no competing interests" statement
% has been inferred. The disclosure must be confirmed by all authors.
\begin{ack}
This work has been supported by the Centre for Spatial Intelligence (RCSI) at University of Bath, the European Union under grant agreement no. 101136006-XTREME, and the European Innovation Council under grant agreement no. 101257536-CEREBRIS.

\noindent\textbf{Competing Interests.}
The authors declare no competing interests.
\end{ack}

\FloatBarrier

{
\small
\bibliographystyle{unsrt}
\bibliography{main}

@String(CVPR= {IEEE Conf. Comput. Vis. Pattern Recog.})

@String(ICCV= {Int. Conf. Comput. Vis.})

@String(ECCV= {Eur. Conf. Comput. Vis.})

@String(ICPR = {Int. Conf. Pattern Recog.})

@String(ICLR = {Int. Conf. Learn. Represent.})

@String(AAAI = {AAAI})

@String(CVPRW= {IEEE Conf. Comput. Vis. Pattern Recog. Worksh.})

@String(CVPR  = {CVPR})

@String(ICCV  = {ICCV})

@String(ECCV  = {ECCV})

@String(ICPR  = {ICPR})

@String(ICLR  = {ICLR})

@String(CVPRW= {CVPRW})

@article{ye2026gigaworldpolicy,
  title={GigaWorld-Policy: An Efficient Action-Centered World--Action Model},
  author={Ye, Angen and Wang, Boyuan and Ni, Chaojun and Huang, Guan and Zhao, Guosheng and Li, Hao and Li, Hengtao and Li, Jie and Lv, Jindi and Liu, Jingyu and Cao, Min and Li, Peng and Deng, Qiuping and Mei, Wenjun and Wang, Xiaofeng and Chen, Xinze and Zhou, Xinyu and Wang, Yang and Chang, Yifan and Li, Yifan and Zhou, Yukun and Ye, Yun and Liu, Zhichao and Zhu, Zheng},
  journal={arXiv preprint arXiv:2603.17240},
  year={2026}
}

@inproceedings{gao2023magicdrive,
  title={{MagicDrive}: Street View Generation with Diverse 3D Geometry Control},
  author={Gao, Ruiyuan and Chen, Kai and Xie, Enze and Hong, Lanqing and Li, Zhenguo and Yeung, Dit-Yan and Xu, Qiang},
  booktitle={International Conference on Learning Representations (ICLR)},
  year={2024}
}

@inproceedings{zhao2025drivedreamer,
  title={Drivedreamer-2: Llm-enhanced world models for diverse driving video generation},
  author={Zhao, Guosheng and Wang, Xiaofeng and Zhu, Zheng and Chen, Xinze and Huang, Guan and Bao, Xiaoyi and Wang, Xingang},
  booktitle={Proceedings of the AAAI Conference on Artificial Intelligence},
  volume={39},
  pages={10412--10420},
  year={2025}
}

@inproceedings{wen2024panacea,
  title={Panacea: Panoramic and controllable video generation for autonomous driving},
  author={Wen, Yuqing and Zhao, Yucheng and Liu, Yingfei and Jia, Fan and Wang, Yanhui and Luo, Chong and Zhang, Chi and Wang, Tiancai and Sun, Xiaoyan and Zhang, Xiangyu},
  booktitle={Proceedings of the IEEE/CVF Conference on Computer Vision and Pattern Recognition},
  pages={6902--6912},
  year={2024}
}

@inproceedings{chen2026memobench,
  title={Memobench: Benchmarking world modeling in dynamically changing environments},
  author={Chen, Haoyu and Zhou, Kaichen and Hua, Hang and Zhang, Kaile and Qian, Jingwen and Ma, Wufei and Chen, Haonan and Liu, Chunjiang and Zhao, Yizhou and Wang, Xiaoyuan and others},
  booktitle={European Conference on Computer Vision},
  pages={658--676},
  year={2026},
  organization={Springer}
}

@article{gao2024vista,
  title={Vista: A generalizable driving world model with high fidelity and versatile controllability},
  author={Gao, Shenyuan and Yang, Jiazhi and Chen, Li and Chitta, Kashyap and Qiu, Yihang and Geiger, Andreas and Zhang, Jun and Li, Hongyang},
  journal={Advances in Neural Information Processing Systems},
  volume={37},
  pages={91560--91596},
  year={2024}
}

@article{tang2026learning,
  title={Learning continuous wasserstein barycenter space for generalized all-in-one image restoration},
  author={Tang, Xiaole and He, Xiaoyi and Xu, Jiayi and Gu, Xiang and Sun, Jian},
  journal={IEEE Transactions on Pattern Analysis and Machine Intelligence},
  year={2026},
  publisher={IEEE}
}

@article{ye2026world,
  title={World Action Models are Zero-shot Policies},
  author={Ye, Seonghyeon and Ge, Yunhao and Zheng, Kaiyuan and Gao, Shenyuan and Yu, Sihyun and Kurian, George and Indupuru, Suneel and Tan, You Liang and Zhu, Chuning and Xiang, Jiannan and Malik, Ayaan and Lee, Kyungmin and Liang, William and Ranawaka, Nadun and Gu, Jiasheng and Xu, Yinzhen and Wang, Guanzhi and Hu, Fengyuan and Narayan, Avnish and Bjorck, Johan and Wang, Jing and Kim, Gwanghyun and Niu, Dantong and Zheng, Ruijie and Xie, Yuqi and Wu, Jimmy and Wang, Qi and Julian, Ryan and Xu, Danfei and Du, Yilun and Chebotar, Yevgen and Reed, Scott and Kautz, Jan and Zhu, Yuke and Fan, {Linxi "Jim"} and Jang, Joel},
  journal={arXiv preprint arXiv:2602.15922},
  year={2026},
  url={https://arxiv.org/abs/2602.15922}
}

@inproceedings{lu2025uni,
  title={Uni-layout: Integrating human feedback in unified layout generation and evaluation},
  author={Lu, Shuo and Chen, Yanyin and Feng, Wei and Fan, Jiahao and Li, Fengheng and Zhang, Zheng and Lv, Jingjing and Shen, Junjie and Law, Ching and Liang, Jian},
  booktitle={Proceedings of the 33rd ACM International Conference on Multimedia},
  pages={7709--7718},
  year={2025}
}

@article{lu2026worldcoder,
  title={WorldCoder-Bench: Benchmarking Physically Grounded 3D World Synthesis},
  author={Lu, Shuo and Xu, Yinuo and Yu, Kecheng and Jiang, Siru and Yu, Yongcan and Wang, Yubin and Yang, Haitao and Zhang, Yuxiang and Wang, Bin and He, Ran and others},
  journal={arXiv preprint arXiv:2606.01869},
  year={2026}
}

@inproceedings{liu2023regformer,
  title={Regformer: An efficient projection-aware transformer network for large-scale point cloud registration},
  author={Liu, Jiuming and Wang, Guangming and Liu, Zhe and Jiang, Chaokang and Pollefeys, Marc and Wang, Hesheng},
  booktitle={2023 IEEE/CVF International Conference on Computer Vision (ICCV)},
  pages={8417--8426},
  year={2023},
  organization={IEEE}
}

@article{liu2025difflow3d,
  title={Difflow3d: Hierarchical diffusion models for uncertainty-aware 3d scene flow estimation},
  author={Liu, Jiuming and Ye, Weicai and Wang, Guangming and Jiang, Chaokang and Pan, Lei and Han, Jinru and Liu, Zhe and Zhang, Guofeng and Wang, Hesheng},
  journal={IEEE transactions on pattern analysis and machine intelligence},
  year={2025},
  publisher={IEEE}
}

@article{liu2026towards,
  title={Towards interactive video world modeling: Frontiers, challenges, benchmarks, and future trends},
  author={Liu, Jiuming and Ni, Chaojun and Liu, Mengmeng and Peng, Chensheng and Wang, Fangjinhua and Shen, Sitian and Pollefeys, Marc and Tomizuka, Masayoshi and Tewari, Ayush and Kristensson, Per Ola},
  journal={arXiv preprint arXiv:2606.01164},
  year={2026}
}

@inproceedings{liu2026arflow,
  title={Arflow: Auto-regressive optical flow estimation for arbitrary-length videos via progressive next-frame forecasting},
  author={Liu, Jiuming and Liu, Mengmeng and Zhu, Siting and Zhang, Yunpeng and Li, Jiangtao and Yang, Michael and Nex, Francesco and Cheng, Hao and Wang, Hesheng},
  booktitle={International Conference on Learning Representations},
  volume={2026},
  pages={130125--130154},
  year={2026}
}

@article{lu2026one,
  title={One size, many fits: Aligning diverse group-wise click preferences in large-scale advertising image generation},
  author={Lu, Shuo and Wang, Haohan and Feng, Wei and Wang, Weizhen and Zhang, Shen and Li, Yaoyu and Ma, Ao and Zhang, Zheng and Lv, Jingjing and Shen, Junjie and others},
  journal={arXiv preprint arXiv:2602.02033},
  year={2026}
}

@INPROCEEDINGS{zhang2025epona,
  author={Zhang, Kaiwen and Tang, Zhenyu and Hu, Xiaotao and Pan, Xingang and Guo, Xiaoyang and Liu, Yuan and Huang, Jingwei and Yuan, Li and Zhang, Qian and Long, Xiao-Xiao and Cao, Xun and Yin, Wei},
  booktitle={2025 IEEE/CVF International Conference on Computer Vision (ICCV)}, 
  title={Epona: Autoregressive Diffusion World Model for Autonomous Driving}, 
  year={2025},
  volume={},
  number={},
  pages={27220-27230},
  doi={10.1109/ICCV51701.2025.02527}}

@inproceedings{shen2025videovla,
  title={{VideoVLA}: Video Generators Can Be Generalizable Robot Manipulators},
  author={Shen, Yichao and Wei, Fangyun and Du, Zhiying and Liang, Yaobo and Lu, Yan and Yang, Jiaolong and Zheng, Nanning and Guo, Baining},
  booktitle={Advances in Neural Information Processing Systems},
  volume={38},
  year={2025},
  doi={10.52202/085713-3197}
}

@inproceedings{caesar2020nuscenes,
  title={nuscenes: A multimodal dataset for autonomous driving},
  author={Caesar, Holger and Bankiti, Varun and Lang, Alex H and Vora, Sourabh and Liong, Venice Erin and Xu, Qiang and Krishnan, Anush and Pan, Yu and Baldan, Giancarlo and Beijbom, Oscar},
  booktitle={Proceedings of the IEEE/CVF conference on computer vision and pattern recognition},
  pages={11621--11631},
  year={2020}
}

@inproceedings{li2025recogdrive,
  title={{ReCogDrive}: A Reinforced Cognitive Framework for End-to-End Autonomous Driving},
  author={Li, Yongkang and Xiong, Kaixin and Guo, Xiangyu and Li, Fang and Yan, Sixu and Xu, Gangwei and Zhou, Lijun and Chen, Long and Sun, Haiyang and Wang, Bing and Ma, Kun and Chen, Guang and Ye, Hangjun and Liu, Wenyu and Wang, Xinggang},
  booktitle={International Conference on Learning Representations (ICLR)},
  year={2026}
}

@inproceedings{hu2023planning,
  title={Planning-oriented autonomous driving},
  author={Hu, Yihan and Yang, Jiazhi and Chen, Li and Li, Keyu and Sima, Chonghao and Zhu, Xizhou and Chai, Siqi and Du, Senyao and Lin, Tianwei and Wang, Wenhai and others},
  booktitle={Proceedings of the IEEE/CVF conference on computer vision and pattern recognition},
  pages={17853--17862},
  year={2023}
}

@inproceedings{weng2024drive,
  title={Para-drive: Parallelized architecture for real-time autonomous driving},
  author={Weng, Xinshuo and Ivanovic, Boris and Wang, Yan and Wang, Yue and Pavone, Marco},
  booktitle={Proceedings of the IEEE/CVF Conference on Computer Vision and Pattern Recognition},
  pages={15449--15458},
  year={2024}
}

@inproceedings{yang2025resim,
  title={{ReSim}: Reliable World Simulation for Autonomous Driving},
  author={Yang, Jiazhi and Chitta, Kashyap and Gao, Shenyuan and Chen, Long and Shao, Yuqian and Jia, Xiaosong and Li, Hongyang and Geiger, Andreas and Yue, Xiangyu and Chen, Li},
  booktitle={Advances in Neural Information Processing Systems},
  volume={38},
  year={2025},
  doi={10.52202/085713-5591}
}

@inproceedings{li2024enhancing,
  title={Enhancing End-to-End Autonomous Driving with Latent World Model},
  author={Li, Yingyan and Fan, Lue and He, Jiawei and Wang, Yuqi and Chen, Yuntao and Zhang, Zhaoxiang and Tan, Tieniu},
  booktitle={International Conference on Learning Representations (ICLR)},
  year={2025}
}

@inproceedings{lu2024wovogen,
  title={Wovogen: World volume-aware diffusion for controllable multi-camera driving scene generation},
  author={Lu, Jiachen and Huang, Ze and Yang, Zeyu and Zhang, Jiahui and Zhang, Li},
  booktitle={European Conference on Computer Vision},
  pages={329--345},
  year={2024},
  organization={Springer}
}

@inproceedings{wang2024drivedreamer,
  title={Drivedreamer: Towards real-world-drive world models for autonomous driving},
  author={Wang, Xiaofeng and Zhu, Zheng and Huang, Guan and Chen, Xinze and Zhu, Jiagang and Lu, Jiwen},
  booktitle={European conference on computer vision},
  pages={55--72},
  year={2024},
  organization={Springer}
}

@inproceedings{wang2024driving,
  title={Driving into the future: Multiview visual forecasting and planning with world model for autonomous driving},
  author={Wang, Yuqi and He, Jiawei and Fan, Lue and Li, Hongxin and Chen, Yuntao and Zhang, Zhaoxiang},
  booktitle={Proceedings of the IEEE/CVF Conference on Computer Vision and Pattern Recognition},
  pages={14749--14759},
  year={2024}
}

@article{hu2023gaia,
  title={Gaia-1: A generative world model for autonomous driving},
  author={Hu, Anthony and Russell, Lloyd and Yeo, Hudson and Murez, Zak and Fedoseev, George and Kendall, Alex and Shotton, Jamie and Corrado, Gianluca},
  journal={arXiv preprint arXiv:2309.17080},
  year={2023}
}

@inproceedings{hu2024drivingworld,
  title={{DrivingWorld}: Constructing World Model for Autonomous Driving via Video {GPT}},
  author={Hu, Xiaotao and Jia, Mingkai and Guo, Xiaoyang and Zhang, Qian and Long, Xiao-xiao and Yin, Wei},
  booktitle={Pattern Recognition (ICPR 2026), Part II},
  series={Lecture Notes in Computer Science},
  volume={16813},
  pages={276--291},
  publisher={Springer},
  address={Cham},
  year={2027},
  doi={10.1007/978-3-032-31583-0_19},
  url={https://link.springer.com/chapter/10.1007/978-3-032-31583-0_19}
}

@article{zheng2024doe,
  title={Doe-1: Closed-loop autonomous driving with large world model},
  author={Zheng, Wenzhao and Xia, Zetian and Huang, Yuanhui and Zuo, Sicheng and Zhou, Jie and Lu, Jiwen},
  journal={arXiv preprint arXiv:2412.09627},
  year={2024}
}

@inproceedings{zhao2025forecasting,
  title={From Forecasting to Planning: Policy World Model for Collaborative State-Action Prediction},
  author={Zhao, Zhida and Fu, Talas and Wang, Yifan and Wang, Lijun and Lu, Huchuan},
  booktitle={Advances in Neural Information Processing Systems},
  volume={38},
  year={2025},
  doi={10.52202/085713-4491}
}

@inproceedings{gao2025magicdrive,
  title={MagicDrive-V2: High-resolution long video generation for autonomous driving with adaptive control},
  author={Gao, Ruiyuan and Chen, Kai and Xiao, Bo and Hong, Lanqing and Li, Zhenguo and Xu, Qiang},
  booktitle={Proceedings of the IEEE/CVF International Conference on Computer Vision},
  pages={28135--28144},
  year={2025}
}

@inproceedings{guo2025genesis,
  title={{Genesis}: Multimodal Driving Scene Generation with Spatio-Temporal and Cross-Modal Consistency},
  author={Guo, Xiangyu and Wu, Zhanqian and Xiong, Kaixin and Xu, Ziyang and Zhou, Lijun and Xu, Gangwei and Xu, Shaoqing and Sun, Haiyang and Wang, Bing and Chen, Guang and Ye, Hangjun and Liu, Wenyu and Wang, Xinggang},
  booktitle={Advances in Neural Information Processing Systems},
  volume={38},
  year={2025},
  doi={10.52202/085713-2021}
}

@inproceedings{li2025omninwm,
  title={{OmniNWM}: Omniscient Driving Navigation World Models},
  author={Li, Bohan and Ma, Zhuang and Du, Dalong and Peng, Baorui and Liang, Zhujin and Liu, Zhenqiang and Guo, Xianda and Zhu, Zheng and Ma, Chao and Jin, Yueming and Jin, Xin and Zhao, Hao and Zeng, Wenjun},
  booktitle={European Conference on Computer Vision},
  series={Lecture Notes in Computer Science},
  volume={17028},
  pages={187--208},
  year={2026},
  organization={Springer},
  doi={10.1007/978-3-032-37624-4_12}
}

@inproceedings{zheng2024occworld,
  title={Occworld: Learning a 3d occupancy world model for autonomous driving},
  author={Zheng, Wenzhao and Chen, Weiliang and Huang, Yuanhui and Zhang, Borui and Duan, Yueqi and Lu, Jiwen},
  booktitle={European conference on computer vision},
  pages={55--72},
  year={2024},
  organization={Springer}
}

@inproceedings{li2025uniscene,
  title={Uniscene: Unified occupancy-centric driving scene generation},
  author={Li, Bohan and Guo, Jiazhe and Liu, Hongsi and Zou, Yingshuang and Ding, Yikang and Chen, Xiwu and Zhu, Hu and Tan, Feiyang and Zhang, Chi and Wang, Tiancai and others},
  booktitle={Proceedings of the Computer Vision and Pattern Recognition Conference},
  pages={11971--11981},
  year={2025}
}

@article{wang2025mila,
  title={MiLA: Multi-view Intensive-fidelity Long-term Video Generation World Model for Autonomous Driving},
  author={Wang, Haiguang and Liu, Daqi and Xie, Hongwei and Liu, Haisong and Ma, Enhui and Yu, Kaicheng and Wang, Limin and Wang, Bing},
  journal={arXiv preprint arXiv:2503.15875},
  year={2025}
}

@article{russell2025gaia,
  title={Gaia-2: A controllable multi-view generative world model for autonomous driving},
  author={Russell, Lloyd and Hu, Anthony and Bertoni, Lorenzo and Fedoseev, George and Shotton, Jamie and Arani, Elahe and Corrado, Gianluca},
  journal={arXiv preprint arXiv:2503.20523},
  year={2025}
}

@article{teed2023deep,
  title={Deep patch visual odometry},
  author={Teed, Zachary and Lipson, Lahav and Deng, Jia},
  journal={Advances in Neural Information Processing Systems},
  volume={36},
  pages={39033--39051},
  year={2023}
}

@article{ji2025cogen,
  title={Cogen: 3d consistent video generation via adaptive conditioning for autonomous driving},
  author={Ji, Yishen and Zhu, Ziyue and Zhu, Zhenxin and Xiong, Kaixin and Lu, Ming and Li, Zhiqi and Zhou, Lijun and Sun, Haiyang and Wang, Bing and Lu, Tong},
  journal={arXiv preprint arXiv:2503.22231},
  year={2025}
}

@inproceedings{min2024driveworld,
  title={Driveworld: 4d pre-trained scene understanding via world models for autonomous driving},
  author={Min, Chen and Zhao, Dawei and Xiao, Liang and Zhao, Jian and Xu, Xinli and Zhu, Zheng and Jin, Lei and Li, Jianshu and Guo, Yulan and Xing, Junliang and others},
  booktitle={Proceedings of the IEEE/CVF conference on computer vision and pattern recognition},
  pages={15522--15533},
  year={2024}
}

@inproceedings{kim2021drivegan,
  title={Drivegan: Towards a controllable high-quality neural simulation},
  author={Kim, Seung Wook and Philion, Jonah and Torralba, Antonio and Fidler, Sanja},
  booktitle={Proceedings of the IEEE/CVF Conference on Computer Vision and Pattern Recognition},
  pages={5820--5829},
  year={2021}
}

@inproceedings{ICLR2026_70925fe2,
 author = {Jiang, Bo and Chen, Shaoyu and Gao, Hao and Liao, Bencheng and Zhang, Qian and Liu, Wenyu and Wang, Xinggang},
 booktitle = {International Conference on Learning Representations},
 editor = {C. Vondrick and B. Hariharan and C. Raffel and L. Pinto and D. Yang and A. Faust},
 pages = {68886--68900},
 title = {VADv2: End-to-End Vectorized Autonomous Driving via Probabilistic Planning},
 url = {https://proceedings.iclr.cc/paper_files/paper/2026/file/70925fe2336ae05917c817d5e4660ab5-Paper-Conference.pdf},
 volume = {2026},
 year = {2026}
}

@article{jia2024bench2drive,
  title={Bench2drive: Towards multi-ability benchmarking of closed-loop end-to-end autonomous driving},
  author={Jia, Xiaosong and Yang, Zhenjie and Li, Qifeng and Zhang, Zhiyuan and Yan, Junchi},
  journal={Advances in Neural Information Processing Systems},
  volume={37},
  pages={819--844},
  year={2024}
}

@inproceedings{jiang2023vad,
  title={Vad: Vectorized scene representation for efficient autonomous driving},
  author={Jiang, Bo and Chen, Shaoyu and Xu, Qing and Liao, Bencheng and Chen, Jiajie and Zhou, Helong and Zhang, Qian and Liu, Wenyu and Huang, Chang and Wang, Xinggang},
  booktitle={Proceedings of the IEEE/CVF International Conference on Computer Vision},
  pages={8340--8350},
  year={2023}
}

@inproceedings{
jang2025dreamgen,
title={DreamGen: Unlocking Generalization in Robot Learning through Video World Models},
author={Joel Jang and Seonghyeon Ye and Zongyu Lin and Jiannan Xiang and Johan Bjorck and Yu Fang and Fengyuan Hu and Spencer Huang and Kaushil Kundalia and Yen-Chen Lin and Lo{\"\i}c Magne and Ajay Mandlekar and Avnish Narayan and You Liang Tan and Guanzhi Wang and Jing Wang and Qi Wang and Yinzhen Xu and Xiaohui Zeng and Kaiyuan Zheng and Ruijie Zheng and Ming-Yu Liu and Luke Zettlemoyer and Dieter Fox and Jan Kautz and Scott Reed and Yuke Zhu and Linxi Fan},
booktitle={9th Annual Conference on Robot Learning},
year={2025},
url={https://openreview.net/forum?id=3CnxNqmklv}
}

@inproceedings{won2025dual,
  title={Dual-Stream Diffusion for World-Model Augmented Vision-Language-Action Model},
  author={Won, John and Lee, Kyungmin and Jang, Huiwon and Kim, Dongyoung and Shin, Jinwoo},
  booktitle={Proceedings of the 43rd International Conference on Machine Learning},
  series={Proceedings of Machine Learning Research},
  volume={306},
  pages={135349--135370},
  publisher={PMLR},
  year={2026}
}

@inproceedings{hu2022stp3,
  title={St-p3: End-to-end vision-based autonomous driving via spatial-temporal feature learning},
  author={Hu, Shengchao and Chen, Li and Wu, Penghao and Li, Hongyang and Yan, Junchi and Tao, Dacheng},
  booktitle={European Conference on Computer Vision},
  pages={533--549},
  year={2022},
  organization={Springer}
}

@inproceedings{zhu2025unified,
  title={Unified World Models: Coupling Video and Action Diffusion for Pretraining on Large Robotic Datasets},
  author={Zhu, Chuning and Yu, Raymond and Feng, Siyuan and Burchfiel, Benjamin and Shah, Paarth and Gupta, Abhishek},
  booktitle={Robotics: Science and Systems (RSS)},
  year={2025}
}

@article{liang2025video,
  title={Video generators are robot policies},
  author={Liang, Junbang and Tokmakov, Pavel and Liu, Ruoshi and Sudhakar, Sruthi and Shah, Paarth and Ambrus, Rares and Vondrick, Carl},
  journal={arXiv preprint arXiv:2508.00795},
  year={2025}
}

@inproceedings{pai2025mimic,
  title={mimic-video: Video-Action Models for Generalizable Robot Control Beyond VLAs},
  author={Pai, Jonas and Achenbach, Liam and Sanchez, Oliver and Charalambous, Stefanos and Montesinos, Victoriano and Forrai, Benedek and Mees, Oier and Nava, Elvis},
  booktitle={Robotics: Science and Systems},
  year={2026},
  doi={10.15607/RSS.2026.XXII.077}
}

@inproceedings{kim2026cosmos,
  title={{Cosmos Policy}: Fine-Tuning Video Models for Visuomotor Control and Planning},
  author={Kim, Moo Jin and Gao, Yihuai and Lin, Tsung-Yi and Lin, Yen-Chen and Ge, Yunhao and Lam, Grace and Liang, Percy and Song, Shuran and Liu, Ming-Yu and Finn, Chelsea and Gu, Jinwei},
  booktitle={International Conference on Learning Representations (ICLR)},
  year={2026}
}

@inproceedings{lingbot-va2026,
  title={Causal World Modeling for Robot Control},
  author={Li, Lin and Zhang, Qihang and Luo, Yiming and Yang, Shuai and Wang, Ruilin and Zhang, Luyao and Yu, Mingrui and Gao, Zelin and Xue, Nan and Zhou, Boyu and Zhu, Xing and Ding, Mingyu and Shen, Yujun and Xu, Yinghao},
  booktitle={Robotics: Science and Systems (RSS)},
  year={2026},
  doi={10.15607/RSS.2026.XXII.016}
}

@article{lecun2022path,
  title={A path towards autonomous machine intelligence version 0.9. 2, 2022-06-27},
  author={LeCun, Yann},
  journal={Open Review},
  volume={62},
  number={1},
  pages={1--62},
  year={2022}
}

@article{hu2023simulation,
  title={How simulation helps autonomous driving: A survey of sim2real, digital twins, and parallel intelligence},
  author={Hu, Xuemin and Li, Shen and Huang, Tingyu and Tang, Bo and Huai, Rouxing and Chen, Long},
  journal={IEEE Transactions on Intelligent Vehicles},
  volume={9},
  number={1},
  pages={593--612},
  year={2023},
  publisher={IEEE}
}

@article{chitta2022transfuser,
  title={Transfuser: Imitation with transformer-based sensor fusion for autonomous driving},
  author={Chitta, Kashyap and Prakash, Aditya and Jaeger, Bernhard and Yu, Zehao and Renz, Katrin and Geiger, Andreas},
  journal={IEEE transactions on pattern analysis and machine intelligence},
  volume={45},
  number={11},
  pages={12878--12895},
  year={2022},
  publisher={IEEE}
}

@inproceedings{tong2023scene,
  title={Scene as occupancy},
  author={Tong, Wenwen and Sima, Chonghao and Wang, Tai and Chen, Li and Wu, Silei and Deng, Hanming and Gu, Yi and Lu, Lewei and Luo, Ping and Lin, Dahua and others},
  booktitle={Proceedings of the IEEE/CVF International Conference on Computer Vision},
  pages={8406--8415},
  year={2023}
}

@inproceedings{zheng2024genad,
  title={Genad: Generative end-to-end autonomous driving},
  author={Zheng, Wenzhao and Song, Ruiqi and Guo, Xianda and Zhang, Chenming and Chen, Long},
  booktitle={European Conference on Computer Vision},
  pages={87--104},
  year={2024},
  organization={Springer}
}

@inproceedings{liao2025diffusiondrive,
  title={Diffusiondrive: Truncated diffusion model for end-to-end autonomous driving},
  author={Liao, Bencheng and Chen, Shaoyu and Yin, Haoran and Jiang, Bo and Wang, Cheng and Yan, Sixu and Zhang, Xinbang and Li, Xiangyu and Zhang, Ying and Zhang, Qian and others},
  booktitle={Proceedings of the Computer Vision and Pattern Recognition Conference},
  pages={12037--12047},
  year={2025}
}

@inproceedings{chen2025drivinggpt,
  title={Drivinggpt: Unifying driving world modeling and planning with multi-modal autoregressive transformers},
  author={Chen, Yuntao and Wang, Yuqi and Zhang, Zhaoxiang},
  booktitle={Proceedings of the IEEE/CVF International Conference on Computer Vision},
  pages={26890--26900},
  year={2025}
}

@article{shi2026honeycomb,
  title={Honeycomb: Constant-Size Scene Memory Representation for Video World Models},
  author={Shi, Jack Wei Lun and Zhou, Kaichen and Chen, Haoyu and Weng, Yufeng and Ong, Keane and Cai, Ruojin and Hua, Hang and Yeoh, Justin KW and Wang, Mengyu},
  journal={arXiv preprint arXiv:2609.37690},
  year={2026}
}

@inproceedings{liu2025dvlo4d,
  title={Dvlo4d: Deep visual-lidar odometry with sparse spatial-temporal fusion},
  author={Liu, Mengmeng and Yang, Michael Ying and Liu, Jiuming and Zhang, Yunpeng and Li, Jiangtao and Oude Elberink, Sander and Vosselman, George and Cheng, Hao},
  booktitle={2025 IEEE International Conference on Robotics and Automation (ICRA)},
  pages={9740--9747},
  year={2025},
  organization={IEEE}
}

@InProceedings{Deng_2026_CVPR,
    author    = {Deng, Tianchen and Chen, Xuefeng and Chen, Yi and Chen, Qu and Xu, Yuyao and Yang, Lijin and Xu, Le and Zhang, Yu and Zhang, Bo and Huang, Wuxiong and Wang, Hesheng},
    title     = {GaussianDWM: 3D Gaussian Driving World Model for Unified Scene Understanding and Multi-Modal Generation},
    booktitle = {Proceedings of the IEEE/CVF Conference on Computer Vision and Pattern Recognition (CVPR)},
    month     = {June},
    year      = {2026},
    pages     = {10656-10667}
}

@article{deng2026gaussiandwm++,
  title={GaussianDWM++: Language-Grounded 3D Gaussian Driving World Model for Unified Scene Understanding, Editing, and Multi-Modal Generation},
  author={Deng, Tianchen and Chen, Xuefeng and Wu, Shuang and Chen, Qu and Zhu, Jiajun and Dai, Bo and Yang, Jianfei and Wang, Hesheng},
  journal={arXiv preprint arXiv:2608.16234},
  year={2026}
}

@article{zhang2026diffusion,
  title={Diffusion Subgoal Planning for Long-Horizon Offline Goal-Conditioned Reinforcement Learning},
  author={Zhang, Hengrui and Cheng, Yuhu and Chen, CL and Wang, Xuesong},
  journal={arXiv preprint arXiv:2609.34575},
  year={2026}
}

@inproceedings{liu20264dstr,
  title={4dstr: Advancing generative 4d gaussians with spatial-temporal rectification for high-quality and consistent 4d generation},
  author={Liu, Mengmeng and Liu, Jiuming and Zhang, Yunpeng and Li, Jiangtao and Yang, Michael Ying and Nex, Francesco and Cheng, Hao},
  booktitle={Proceedings of the AAAI Conference on Artificial Intelligence},
  volume={40},
  pages={7224--7232},
  year={2026}
}

@inproceedings{liu2023tracing,
  title={Tracing the influence of predecessors on trajectory prediction},
  author={Liu, Mengmeng and Cheng, Hao and Yang, Michael Ying},
  booktitle={2023 IEEE/CVF International Conference on Computer Vision Workshops (ICCVW)},
  pages={3245--3255},
  year={2023},
  organization={IEEE}
}

@inproceedings{liu2026streamvlo,
  title={Streamvlo: Streaming visual-lidar odometry with cumulative drift compensation},
  author={Liu, Mengmeng and Liu, Jiuming and Yang, Michael Ying and Jiang, Chaokang and Li, Jiangtao and Zhang, Yunpeng and Wang, Hesheng and Nex, Francesco and Cheng, Hao},
  booktitle={Proceedings of the IEEE/CVF Conference on Computer Vision and Pattern Recognition},
  pages={39086--39097},
  year={2026}
}

@article{cheng2023gatraj,
  title={Gatraj: A graph-and attention-based multi-agent trajectory prediction model},
  author={Cheng, Hao and Liu, Mengmeng and Chen, Lin and Broszio, Hellward and Sester, Monika and Yang, Michael Ying},
  journal={ISPRS Journal of Photogrammetry and Remote Sensing},
  volume={205},
  pages={163--175},
  year={2023},
  publisher={Elsevier}
}

@inproceedings{liu2024laformer,
  title={Laformer: Trajectory prediction for autonomous driving with lane-aware scene constraints},
  author={Liu, Mengmeng and Cheng, Hao and Chen, Lin and Broszio, Hellward and Li, Jiangtao and Zhao, Runjiang and Sester, Monika and Yang, Michael Ying},
  booktitle={2024 IEEE/CVF Conference on Computer Vision and Pattern Recognition Workshops (CVPRW)},
  pages={2039--2049},
  year={2024},
  organization={IEEE}
}

@inproceedings{liu2026driveva,
  title={{DriveVA}: Video Action Models are Zero-Shot Drivers},
  author={Liu, Mengmeng and Zhang, Diankun and Liu, Jiuming and Cui, Jianfeng and Xie, Hongwei and Chen, Guang and Ye, Hangjun and Yang, Michael Ying and Nex, Francesco and Cheng, Hao},
  booktitle={Computer Vision -- ECCV 2026},
  series={Lecture Notes in Computer Science},
  volume={17066},
  pages={315--335},
  publisher={Springer},
  year={2026},
  doi={10.1007/978-3-032-37718-0_19}
}

@article{li2025unified,
  title={Unified video action model},
  author={Li, Shuang and Gao, Yihuai and Sadigh, Dorsa and Song, Shuran},
  journal={arXiv preprint arXiv:2503.00200},
  year={2025}
}

@article{ha2018world,
  title={World models},
  author={Ha, David and Schmidhuber, J{\"u}rgen},
  journal={arXiv preprint arXiv:1803.10122},
  year={2018}
}

@inproceedings{peebles2023DIT,
  title={Scalable diffusion models with transformers},
  author={Peebles, William and Xie, Saining},
  booktitle={Proceedings of the IEEE/CVF international conference on computer vision},
  pages={4195--4205},
  year={2023}
}

@article{wan2025,
      title={Wan: Open and Advanced Large-Scale Video Generative Models}, 
      author={Team Wan and Ang Wang and Baole Ai and Bin Wen and Chaojie Mao and Chen-Wei Xie and Di Chen and others},
      journal = {arXiv preprint arXiv:2503.20314},
      year={2025}
}

@article{liu2026palm,
  title={Palm: Progress-aware policy learning via affordance reasoning for long-horizon robotic manipulation},
  author={Liu, Yuanzhe and Zhu, Jingyuan and Mo, Yuchen and Li, Gen and Cao, Xu and Jin, Jin and Shen, Yifan and Li, Zhengyuan and Yu, Tianjiao and Yuan, Wenzhen and others},
  journal={arXiv preprint arXiv:2601.07060},
  year={2026}
}

@article{shen2026fine,
  title={Fine-grained preference optimization improves spatial reasoning in vlms},
  author={Shen, Yifan and Liu, Yuanzhe and Zhu, Jingyuan and Cao, Xu and Zhang, Xiaofeng and He, Yixiao and Ye, Wenming and Rehg, James and Lourentzou, Ismini},
  journal={Advances in Neural Information Processing Systems},
  volume={38},
  pages={17929--17960},
  year={2025}
}

@article{meng2026phymetric,
  title={PhyMetric: Diagnosing Physical Plausibility in Text-to-Video Generation via Scene-Level QA},
  author={Meng, Siwei and Cai, Xuerou and Hu, Rui and Liu, Ping},
  journal={arXiv preprint arXiv:2610.19823},
  year={2026}
}

@article{yu2026elsa3d,
  title={ELSA3D: Elastic Semantic Anchoring for Unified 3D Understanding and Generation},
  author={Yu, Tianjiao and Li, Xinzhuo and Shen, Yifan and Susladkar, Onkar and Liu, Yuanzhe and Zhou, Xiaona and Lourentzou, Ismini},
  journal={arXiv preprint arXiv:2607.06565},
  year={2026}
}

@article{dong2026language,
  title={Language-Conditioned World Modeling for Visual Navigation},
  author={Dong, Yifei and Wu, Fengyi and Dai, Yilong and Kong, Lingdong and Chen, Guangyu and Sha, Yetong and Hu, Qiyu and Liu, Feng and Huang, Siyu and Dai, Qi and Cheng, Zhi-Qi},
  journal={arXiv preprint arXiv:2603.26741},
  year={2026}
}

@article{xie2026step,
  title={Step-wise rubric rewards for llm reasoning},
  author={Xie, Weichu and Zhao, Haozhe and Liu, Wenpu and Zhu, Yongfu and Chen, Liang and Ye, Minghao and Chen, Zirong and Xu, Yuqi and Dong, Shuai and Wang, Ziyue and others},
  journal={arXiv preprint arXiv:2605.17291},
  year={2026}
}

@inproceedings{dong2026towards,
  title={Towards Unified World Models for Visual Navigation via Memory-Augmented Planning and Foresight},
  author={Dong, Yifei and Wu, Fengyi and Chen, Guangyu and Kong, Lingdong and Hu, Qiyu and Zhou, Yuxuan and Zhu, Xu and Sun, Jingdong and He, Jun-Yan and Dai, Qi and Hauptmann, Alexander G. and Cheng, Zhi-Qi},
  booktitle={European Conference on Computer Vision},
  pages={276--295},
  year={2026},
  organization={Springer}
}

@article{dong2025ha,
  title={{HA-VLN} 2.0: An Open Benchmark and Leaderboard for Human-Aware Navigation in Discrete and Continuous Environments with Dynamic Multi-Human Interactions},
  author={Dong, Yifei and Wu, Fengyi and He, Qi and Kong, Lingdong and Li, Heng and Li, Minghan and Cheng, Zebang and Zhou, Yuxuan and Sun, Jingdong and Dai, Qi and Hauptmann, Alexander G. and Cheng, Zhi-Qi},
  journal={arXiv preprint arXiv:2503.14229},
  year={2025}
}

@INPROCEEDINGS{li2025end_wote,
  author={Li, Yingyan and Wang, Yuqi and Liu, Yang and He, Jiawei and Fan, Lue and Zhang, Zhaoxiang},
  booktitle={2025 IEEE/CVF International Conference on Computer Vision (ICCV)}, 
  title={End-to-End Driving with Online Trajectory Evaluation via BEV World Model}, 
  year={2025},
  volume={},
  number={},
  pages={27137-27146},
  doi={10.1109/ICCV51701.2025.02519}}

@article{wang2025prophetdwm,
  title={Prophetdwm: A driving world model for rolling out future actions and videos},
  author={Wang, Xiaodong and Peng, Peixi},
  journal={arXiv preprint arXiv:2505.18650},
  year={2025}
}

@article{li2025imagidrive,
  title={{ImagiDrive}: A Unified Imagination-and-Planning Framework for Autonomous Driving},
  author={Li, Jingyu and Zhang, Bozhou and Jin, Xin and Deng, Jiankang and Zhu, Xiatian and Zhang, Li},
  journal={arXiv preprint arXiv:2508.11428},
  year={2025},
  url={https://arxiv.org/abs/2508.11428}
}

@inproceedings{NUSCENES,
  title={nuscenes: A multimodal dataset for autonomous driving},
  author={Caesar, Holger and Bankiti, Varun and Lang, Alex H and Vora, Sourabh and Liong, Venice Erin and Xu, Qiang and Krishnan, Anush and Pan, Yu and Baldan, Giancarlo and Beijbom, Oscar},
  booktitle={Proceedings of the IEEE/CVF conference on computer vision and pattern recognition},
  pages={11621--11631},
  year={2020}
}

@article{dauner2024navsim,
  title={Navsim: Data-driven non-reactive autonomous vehicle simulation and benchmarking},
  author={Dauner, Daniel and Hallgarten, Marcel and Li, Tianyu and Weng, Xinshuo and Huang, Zhiyu and Yang, Zetong and Li, Hongyang and Gilitschenski, Igor and Ivanovic, Boris and Pavone, Marco and others},
  journal={Advances in Neural Information Processing Systems},
  volume={37},
  pages={28706--28719},
  year={2024}
}

@inproceedings{dosovitskiy2017carla,
  title={CARLA: An open urban driving simulator},
  author={Dosovitskiy, Alexey and Ros, German and Codevilla, Felipe and Lopez, Antonio and Koltun, Vladlen},
  booktitle={Conference on robot learning},
  pages={1--16},
  year={2017},
  organization={PMLR}
}

@misc{openscene2023,
  title={OpenScene: The Largest Up-to-Date 3D Occupancy Prediction Benchmark in Autonomous Driving},
  author={OpenScene Contributors},
  howpublished={\url{https://github.com/OpenDriveLab/OpenScene}},
  year={2023}
}

@InProceedings{xia2025drivelaw,
    author    = {Xia, Tianze and Li, Yongkang and Zhou, Lijun and Yao, Jingfeng and Xiong, Kaixin and Sun, Haiyang and Wang, Bing and Ma, Kun and Chen, Guang and Ye, Hangjun and Liu, Wenyu and Wang, Xinggang},
    title     = {DriveLaW: Unifying Planning and Video Generation in a Latent Driving World},
    booktitle = {Proceedings of the IEEE/CVF Conference on Computer Vision and Pattern Recognition (CVPR)},
    month     = {June},
    year      = {2026},
    pages     = {39701-39712}
}

@inproceedings{li2025drivevlaw0,
  title={{DriveVLA-W0}: World Models Amplify Data Scaling Law in Autonomous Driving},
  author={Li, Yingyan and Shang, Shuyao and Liu, Weisong and Zhan, Bing and Wang, Haochen and Wang, Yuqi and Chen, Yuntao and Wang, Xiaoman and An, Yasong and Tang, Chufeng and Hou, Lu and Fan, Lue and Zhang, Zhaoxiang},
  booktitle={International Conference on Learning Representations (ICLR)},
  year={2026}
}

@article{yuan2026fast,
  title={Fast-WAM: Do World Action Models Need Test-time Future Imagination?},
  author={Yuan, Tianyuan and Dong, Zibin and Liu, Yicheng and Zhao, Hang},
  journal={arXiv preprint arXiv:2603.16666},
  year={2026}
}
}

%%%%%%%%%%%%%%%%%%%%%%%%%%%%%%%%%%%%%%%%%%%%%%%%%%%%%%%%%%%%

% Continue the appendix after the references to avoid a single-reference page.
\par\bigskip
\appendix

\section{Supplementary Material}
\label{sec:supp}

This appendix provides architecture and inference details (Sec.~\ref{sec:supp_architecture}), additional ablations and generalization analysis (Sec.~\ref{sec:supp_additional_experiments}), and the zero-shot evaluation protocol (Sec.~\ref{sec:supp_protocol}).
It also retains DPVO verification (Sec.~\ref{sec:supp_dpvo}), the extended nuScenes comparison (Sec.~\ref{sec:supp_nuscenes}), qualitative results (Sec.~\ref{sec:supp_visual}), limitations, asset licenses, and broader impacts.

\section{Additional Architecture Details}
\label{sec:supp_architecture}

\subsection{Visibility Mask Implementation}
\label{sec:supp_mask}

The main paper introduces the Modality-Decoupling Visibility Mask conceptually.
Here we provide its explicit block form.
Let the self-attention token groups be historical context tokens $C$, future video tokens $V$, and future trajectory tokens $A$.
The binary visibility mask is
\begin{equation}
\mathbf{M} =
\begin{array}{c|ccc}
      & C & V & A \\
\hline
C     & 1 & 0 & 0 \\
V     & 1 & 1 & 0 \\
A     & 1 & 0 & 1
\end{array},
\end{equation}
where rows denote query tokens, columns denote key/value tokens, $1$ means attention is allowed, and $0$ means attention is blocked.

This mask enforces three constraints.
First, both future modalities read the same historical context through $V\!\rightarrow\!C$ and $A\!\rightarrow\!C$ attention.
Second, context tokens remain read-only with respect to future targets, since $C\!\rightarrow\!V$ and $C\!\rightarrow\!A$ are blocked.
This prevents context tokens from acting as an indirect information bridge between future video and trajectory tokens.
Third, future video and trajectory tokens are mutually invisible through blocked $V\!\rightarrow\!A$ and $A\!\rightarrow\!V$ attention.
Thus, video and trajectory denoising share model parameters but remain mutually invisible at the future-token level.

Full intra-modality attention is preserved within $V$ and within $A$.
This is valid because \textsc{UNIVERSE} predicts future video and trajectory chunks with diffusion-style denoising rather than autoregressive next-step decoding.
Bidirectional attention within $V$ helps maintain temporal coherence across generated frames, while bidirectional attention within $A$ helps produce smooth trajectory chunks.

\subsection{Inference Token Configurations}
\label{sec:supp_inference}

At inference time, \textsc{UNIVERSE} uses the same trained DiT weights but instantiates different future token groups.
Table~\ref{tab:supp_inference_modes} summarizes the token configurations for each inference mode.

\begin{table}[h]
\centering
\caption{\textbf{Inference token configurations.}
Different inference modes are obtained by changing the instantiated future token groups while keeping the same shared DiT weights.}
\label{tab:supp_inference_modes}
\small
\setlength{\tabcolsep}{4pt}
\resizebox{0.95\linewidth}{!}{
\begin{tabular}{l|ccc}
\toprule
\textbf{Mode} & \textbf{Context $C$} & \textbf{Future Video $V$} & \textbf{Future Trajectory $A$} \\
\midrule
Trajectory-only rollout & \cmark & -- & \cmark \\
Video-only rollout & \cmark & \cmark & -- \\
Joint video-action rollout & \cmark & \cmark & \cmark \\
\bottomrule
\end{tabular}}
\end{table}

The key property is dependency consistency.
During training, trajectory tokens are blocked from attending to future video tokens.
Therefore, removing $V$ during trajectory-only inference does not change the attention dependency of the trajectory branch.
This differs from naively coupled video-action generation, where removing future video tokens at test time would alter the learned dependency structure and introduce a train--test mismatch.

% Quantitative values in this file are transcribed from the uploaded OpenReview
% author responses (28 July and 4 August 2026). No new experiments were run.
% Source-page mapping is provided in CAMERA_READY_CHANGES.md, not in the paper.

\section{Additional Experimental Analyses}
\label{sec:supp_additional_experiments}

% \subsection{Comparison of Pretrained Video Backbones}
% \label{sec:supp_backbones}

% Table~\ref{tab:supp_backbones} compares Cosmos-Predict2.5-2B~\citep{nvidia2025cosmospredict25card}, Cosmos3-Edge (4B)~\citep{nvidia2026cosmos3edgecard}, and the default Wan2.2-TI2V-5B~\citep{wan2025}, with $\lambda_{\mathrm{vid}}=1.0$.
% The comparison varies the pretrained backbone while retaining the video-action formulation and loss weight.
% All three configurations yield strong NAVSIM and zero-shot results, but the comparison changes model family and pretraining as well as size; it is not a pure parameter-scaling experiment.
% The sizes in the table identify the nominal backbone configurations rather than an audited count of every auxiliary component.

% \begin{table}[htbp]
% \centering
% \caption{\textbf{Pretrained backbone comparison.} The video-loss weight is fixed to $1.0$. $\dagger$ marks the default. L2 is in meters and collision is in percent.}
% \label{tab:supp_backbones}
% \small
% \setlength{\tabcolsep}{4pt}
% \begin{tabular}{@{}lcccccc@{}}
% \toprule
% Backbone & Size & NAVSIM & \multicolumn{2}{c}{nuScenes} & \multicolumn{2}{c}{Bench2Drive} \\
% \cmidrule(lr){3-3}\cmidrule(lr){4-5}\cmidrule(l){6-7}
%  & & PDMS$\uparrow$ & L2$\downarrow$ & Col.$\downarrow$ & L2$\downarrow$ & Col.$\downarrow$ \\
% \midrule
% Cosmos-Predict2.5 & 2B & 90.7 & 0.85 & \textbf{0.03} & 1.37 & \textbf{1.18} \\
% Cosmos3-Edge & 4B & 90.8 & 0.86 & 0.04 & 1.35 & 1.21 \\
% Wan2.2-TI2V$^\dagger$ & 5B & \textbf{91.0} & \textbf{0.83} & \textbf{0.03} & \textbf{1.33} & 1.19 \\
% \bottomrule
% \end{tabular}
% \end{table}

\subsection{Source-Domain Fitting and Generalization}
\label{sec:supp_generalization}

By \emph{dataset-specific action priors}, we mean frequent source-dataset motion patterns that may help predict ego actions without fully using scene-dependent dynamics.
The same high-level command can require continuing, braking, bypassing, or yielding depending on surrounding agents and obstacles.
We examine source-domain fitting and NAVSIM Navtest performance using the same evaluation protocol (Table~\ref{tab:supp_generalization}).
Two-DiT and UNIVERSE have nearly identical training PDMS, but UNIVERSE has a smaller Train--Navtest gap and stronger zero-shot performance (Table~\ref{tab:zero_shot}).
Thus, the weaker NAVSIM Navtest performance of Two-DiT is not accompanied by weaker training-set fit in this comparison.

\begin{table}[htbp]
\centering
\caption{\textbf{Source-domain fitting and NAVSIM Train--Navtest gap.} Gap is NAVSIM Train PDMS minus NAVSIM Navtest PDMS, in score points.}
\label{tab:supp_generalization}
\small
\setlength{\tabcolsep}{7pt}
\begin{tabular}{@{}lccc@{}}
\toprule
Method & Train PDMS$\uparrow$ & Navtest PDMS$\uparrow$ & Train--Navtest Gap$\downarrow$ \\
\midrule
Two-DiT & 93.78 & 89.6 & 4.18 \\
\ourmethod & 93.80 & \textbf{91.0} & \textbf{2.80} \\
\bottomrule
\end{tabular}
\end{table}

The average DPVO-based consistency error also decreases from $0.19$ for Two-DiT to $0.13$ for UNIVERSE (Table~\ref{tab:dpvo_consistency}).
Together, these observations are consistent with a more transferable representation under shared video-action training.
They are supportive rather than conclusive evidence about action-prior reliance: neither a Train--Navtest gap nor zero-shot collision rate uniquely measures this mechanism, and we do not claim a causal decomposition of the improvement.

\subsection{Submetric Trade-offs and Inference Modes}
\label{sec:supp_tradeoffs}

The objective optimizes trajectory and video flow matching, not individual NAVSIM submetrics.
Accordingly, small trajectory changes can trade off NC, DAC, TTC, comfort, and progress; the full model need not dominate every submetric.
The no-mask variant has slightly higher NC and TTC, whereas UNIVERSE has higher aggregate PDMS, lower reported cross-domain collision rates, and substantially lower trajectory-inference latency.
The visibility mask adds no learned parameters and resolves the tension between shared training parameters and independently selectable outputs.

Trajectory-only inference is the mode used for the principal planning and efficiency results.
Video-only and joint inference demonstrate flexible output capabilities, and joint outputs support the reported consistency analyses.
Synthetic-data generation, decision diagnosis, and self-simulation are potential uses, not separately validated downstream improvements in this work.

\subsection{Additional Ablation on Horizon and Sampling}
\label{sec:horizon_steps}

\begin{table}[htbp]
\centering
\caption{\textbf{Ablation study on future video frames and sampling steps.}
We use $8$ future frames and $2$ sampling steps as the default setting.}
\label{tab:ablation}
\small
\setlength{\tabcolsep}{5pt}
\begin{tabular}{@{}llcccccc@{}}
\toprule
Factor & Value & NC$\uparrow$ & DAC$\uparrow$ & TTC$\uparrow$ & Comf.$\uparrow$ & EP$\uparrow$ & PDMS$\uparrow$ \\
\midrule
Future frames & 4 & 98.4 & 92.3 & 97.8 & 99.5 & 77.6 & 84.9 \\
& 8 & \textbf{99.1} & \textbf{97.6} & \textbf{98.5} & \textbf{100} & \textbf{83.6} & \textbf{91.0} \\
& 12 & 98.8 & 94.8 & 98.3 & \textbf{100} & 80.8 & 88.1 \\
\midrule
Sampling steps & 1 & 61.2 & 37.9 & 50.4 & 1.8 & 14.1 & 13.7 \\
& 2 & \textbf{99.1} & \textbf{97.6} & 98.5 & \textbf{100} & \textbf{83.6} & \textbf{91.0} \\
& 3 & \textbf{99.1} & \textbf{97.6} & \textbf{98.6} & \textbf{100} & 83.5 & \textbf{91.0} \\
\bottomrule
\end{tabular}
\end{table}

Table~\ref{tab:ablation} studies the number of future video frames and flow sampling steps.
Using $4$ future frames provides insufficient long-horizon supervision, while $8$ frames achieves the best PDMS of $91.0$.
Further increasing the horizon to $12$ frames brings no additional gain.
For sampling, one step is unstable, whereas two steps already recover full performance and three steps offer negligible improvement.
We therefore use $8$ future frames and $2$ sampling steps by default.

\FloatBarrier

\section{Zero-Shot Evaluation Protocol}
\label{sec:supp_protocol}

\subsection{Data Splits, Inputs, and Preprocessing}
\label{sec:supp_zero_shot_protocol}

All cross-dataset results use the same checkpoint trained exclusively on NAVSIM v1, without target-domain fine-tuning or adaptation.
Table~\ref{tab:supp_zero_shot_protocol} summarizes the settings used for the open-loop evaluations.
The eight predicted trajectory points cover $4$\,s, while the main comparison reports errors at $1$, $2$, and $3$\,s and the reported averages.

\begin{table}[htbp]
\centering
\caption{\textbf{Open-loop zero-shot evaluation settings.} Both datasets use front-camera history only, together with current ego velocity and a high-level command.}
\label{tab:supp_zero_shot_protocol}
\small
\setlength{\tabcolsep}{4pt}
\renewcommand{\arraystretch}{1.15}
\begin{tabularx}{\linewidth}{@{}p{0.22\linewidth}XX@{}}
\toprule
Setting & nuScenes & Bench2Drive \\
\midrule
Evaluation split & Official v1.0-trainval/val split (150 scenes) & Official validation set \\
Camera & CAM\_FRONT only & CAM\_FRONT only \\
Temporal sampling & 2 Hz & Downsampled from 10 Hz to 2 Hz \\
History & Four frames & Four frames \\
Prediction & Eight trajectory points over 4 s & Eight trajectory points over 4 s \\
Resolution & $832\times480$ & $832\times480$ \\
Checkpoint & \multicolumn{2}{c}{The same NAVSIM-trained checkpoint; no target adaptation} \\
\bottomrule
\end{tabularx}
\end{table}

\noindent\textbf{Preprocessing.}
Images are converted to RGB, resized to $832\times480$, and normalized to $[-1,1]$, without test-time cropping, augmentation, undistortion, or multi-view stitching.
Trajectories are converted to a unified ego-centric coordinate system centered at the current ego pose; nuScenes global poses and Bench2Drive native poses are transformed accordingly.

\noindent\textbf{Conditioning.}
The policy receives front-camera history, current ego velocity, and a left/straight/right high-level command.
Its inputs do not include camera intrinsics/extrinsics, future coordinates, maps, bounding boxes, depth, occupancy, or LiDAR.
This concerns the policy's conditioning inputs, not the benchmark annotations used to compute evaluation metrics.

\noindent\textbf{No target-domain adaptation.}
There is no target-domain fine-tuning, LoRA, test-time optimization, parameter calibration, checkpoint selection, or normalization fitting.

\noindent\textbf{Model availability.}
The released UNIVERSE checkpoint is available at
\url{https://huggingface.co/mengmengliu1998/UNIVERSE}.
The preceding protocol specifies the evaluation splits, conditioning inputs, preprocessing, and absence of target-domain adaptation; public benchmark assets and external dependencies remain subject to their providers' respective terms.

\FloatBarrier

\section{DPVO-based Verification of Video-Action Consistency}
\label{sec:supp_dpvo}

We use DPVO~\citep{teed2023deep} as an external visual-odometry system to verify whether generated videos and predicted trajectories describe consistent ego motion.
For each sample, DPVO is run on both the ground-truth future video and the future video generated by \textsc{UNIVERSE}.
This produces two reconstructed trajectories, denoted as DPVO(gt img) and DPVO(pred img).
We compare DPVO(gt img) with the ground-truth ego trajectory, and DPVO(pred img) with the trajectory predicted by \textsc{UNIVERSE}.

Since monocular visual odometry is ambiguous up to scale, rotation, and translation, we align each reconstructed trajectory to its reference using a 2D similarity transform.
Let $\tau=\{\mathbf{p}_t\}_{t=1}^{T}$ be the reference trajectory and $\hat{\tau}=\{\hat{\mathbf{p}}_t\}_{t=1}^{T}$ be the DPVO-reconstructed trajectory on the ground plane.
We solve
\begin{equation}
    (s^\star,R^\star,\mathbf{t}^\star)
    =
    \arg\min_{s,R,\mathbf{t}}
    \sum_{t=1}^{T}
    \left\|
    sR\hat{\mathbf{p}}_t+\mathbf{t}-\mathbf{p}_t
    \right\|_2^2 ,
\end{equation}
and compute the aligned trajectory as
$\tilde{\mathbf{p}}_t=s^\star R^\star\hat{\mathbf{p}}_t+\mathbf{t}^\star$.
The final consistency error is the average L2 distance over the future 4s horizon:
\begin{equation}
\mathrm{Avg.\ L2}(\tilde{\tau},\tau)=
\frac{1}{T}\sum_{t=1}^{T}
\left\|\tilde{\mathbf{p}}_t-\mathbf{p}_t\right\|_2 .
\label{eq:supp_avg_l2_consistency}
\end{equation}

\begin{figure}[htbp]
    \centering
    \includegraphics[width=1.0\linewidth]{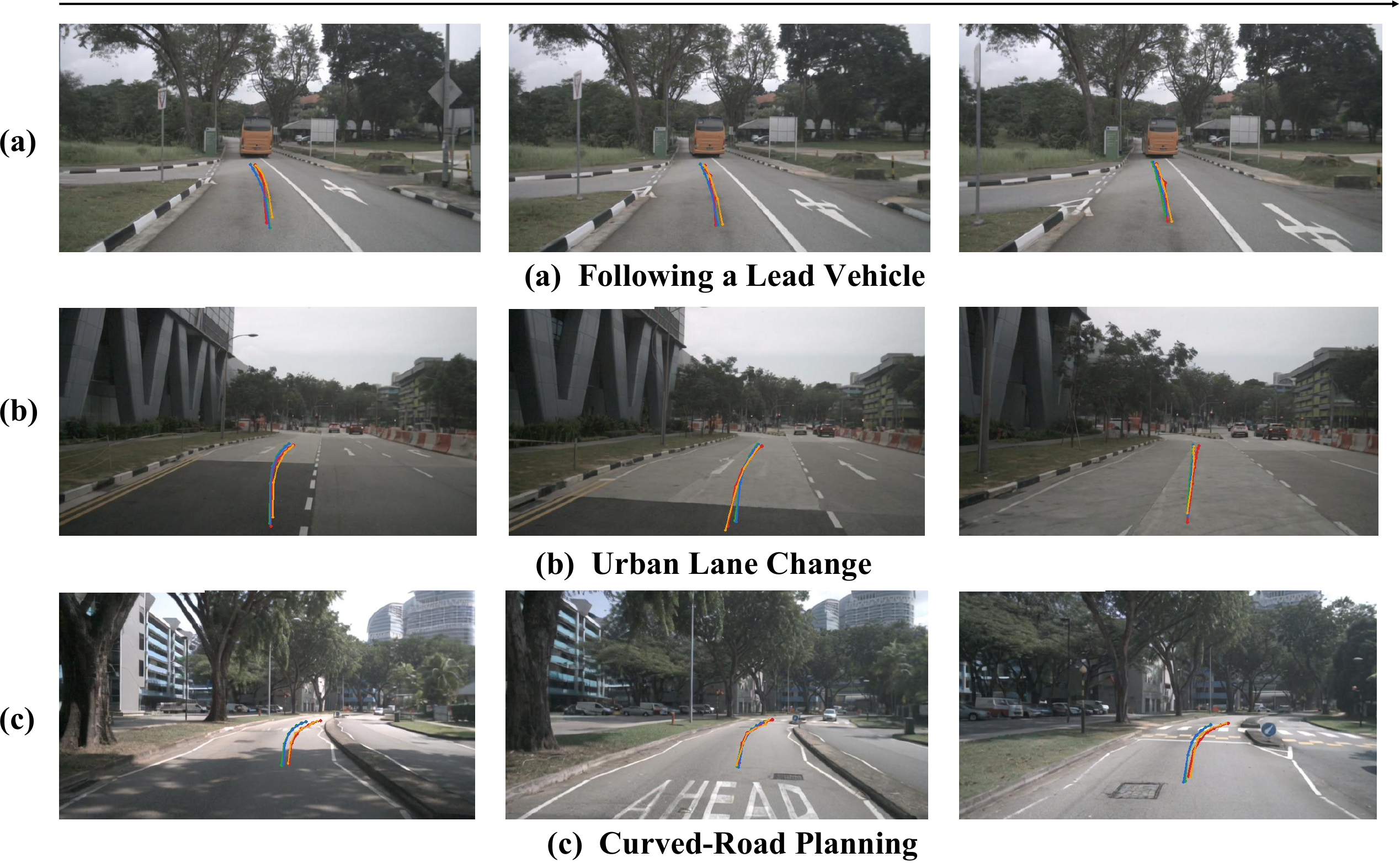}
    \caption{\textbf{Zero-shot video-action consistency across diverse driving scenarios.}
    \textsc{UNIVERSE} is directly evaluated on unseen nuScenes and Bench2Drive scenes without fine-tuning, covering lead-vehicle following, urban lane change, and curved-road planning.
    For nuScenes, DPVO~\citep{teed2023deep} is used to reconstruct ego trajectories from both ground-truth and generated future videos.
    \textcolor{green}{GT Future} and \textcolor{red}{Pred Future} denote ground-truth and predicted trajectories, while \textcolor{dpvogat}{DPVO(gt img)} and \textcolor{dpvopred}{DPVO(pred img)} denote DPVO reconstructions from ground-truth and predicted videos.
    The close alignment among these trajectories shows that the generated visual dynamics remain consistent with the predicted ego motion under domain shift.}
    \label{fig:supp_dpvo}
    \vspace{-0.25cm}
\end{figure}

As reported in Table~\ref{tab:dpvo_consistency}, \textsc{UNIVERSE} achieves an average DPVO-based error of $0.13$, close to the GT-reference error of $0.08$.
This indicates that the ego motion implied by the generated future videos is consistent with the trajectory predicted by the model.
Fig.~\ref{fig:supp_dpvo} further shows that DPVO(pred img) closely follows the predicted trajectory across diverse zero-shot driving scenarios, supporting the video-action consistency of the unified generative process.

\section{Additional Comparison on nuScenes}
\label{sec:supp_nuscenes}

Table~\ref{tab:supp_plan_nusc} provides an extended comparison on the nuScenes planning benchmark.
Different from Table~\ref{tab:zero_shot} in the main paper, which focuses on zero-shot transfer from NAVSIM to nuScenes and Bench2Drive, this table additionally includes methods trained or fine-tuned on nuScenes.
Although \textsc{UNIVERSE} is directly evaluated on nuScenes without any fine-tuning, it achieves lower L2 error and collision rate than the listed nuScenes-trained or fine-tuned baselines.
This further suggests that unified video-action co-training learns transferable planning priors instead of only fitting the source-domain trajectory distribution.

\begin{table*}[htbp]
\centering
\caption{\textbf{End-to-end motion planning performance on the nuScenes~\citep{caesar2020nuscenes} dataset.}
$^*$ indicates that only the front camera is used as input.
Although \textsc{UNIVERSE} is not fine-tuned on nuScenes, it outperforms prior methods trained or fine-tuned on nuScenes.}
\label{tab:supp_plan_nusc}
\vspace{-0.2cm}
\setlength{\tabcolsep}{3pt}
\resizebox{0.98\linewidth}{!}{
\begin{tabular}{l|cc|lc|cccc|cccc}
\toprule
\multirow{2}{*}{Method} &
\multirow{2}{*}{\shortstack{nuScenes\\Finetune}} &
\multirow{2}{*}{Ref} &
\multirow{2}{*}{Input} &
\multirow{2}{*}{Auxiliary Supervision} &
\multicolumn{4}{c|}{L2 (m) $\downarrow$} &
\multicolumn{4}{c}{Collision Rate (\%) $\downarrow$} \\
&&&&& 1s & 2s & 3s & \cellcolor{gray!30}Avg. & 1s & 2s & 3s & \cellcolor{gray!30}Avg. \\
\midrule
ST-P3~\citep{hu2022stp3}
& \cmark & ECCV'22 & Camera & Map\&Box\&Depth
& 1.33 & 2.11 & 2.90 & \cellcolor{gray!30}2.11
& 0.23 & 0.62 & 1.27 & \cellcolor{gray!30}0.71 \\
UniAD~\citep{hu2023planning}
& \cmark & CVPR'23 & Camera & {\footnotesize Map\&Box\&Motion}
& 0.48 & 0.96 & 1.65 & \cellcolor{gray!30}1.03
& 0.05 & 0.17 & 0.71 & \cellcolor{gray!30}0.31 \\
OccNet~\citep{tong2023scene}
& \cmark & ICCV'23 & Camera & 3D-Occ\&Map\&Box
& 1.29 & 2.13 & 2.99 & \cellcolor{gray!30}2.14
& 0.21 & 0.59 & 1.37 & \cellcolor{gray!30}0.72 \\
OccWorld~\citep{zheng2024occworld}
& \cmark & ECCV'24 & Camera & 3D-Occ
& 0.52 & 1.27 & 2.41 & \cellcolor{gray!30}1.40
& 0.12 & 0.40 & 2.08 & \cellcolor{gray!30}0.87 \\
VAD-Tiny~\citep{jiang2023vad}
& \cmark & ICCV'23 & Camera & Map\&Box\&Motion
& 0.60 & 1.23 & 2.06 & \cellcolor{gray!30}1.30
& 0.31 & 0.53 & 1.33 & \cellcolor{gray!30}0.72 \\
VAD-Base~\citep{jiang2023vad}
& \cmark & ICCV'23 & Camera & Map\&Box\&Motion
& 0.54 & 1.15 & 1.98 & \cellcolor{gray!30}1.22
& 0.04 & 0.39 & 1.17 & \cellcolor{gray!30}0.53 \\
GenAD~\citep{zheng2024genad}
& \cmark & ECCV'24 & Camera & Map\&Box\&Motion
& 0.36 & 0.83 & 1.55 & \cellcolor{gray!30}0.91
& 0.06 & 0.23 & 1.00 & \cellcolor{gray!30}0.43 \\
\midrule
Doe-1~\citep{zheng2024doe}
& \cmark & arXiv'24 & Camera$^*$ & QA
& 0.50 & 1.18 & 2.11 & \cellcolor{gray!30}1.26
& 0.04 & 0.37 & 1.19 & \cellcolor{gray!30}0.53 \\
Epona~\citep{zhang2025epona}
& \cmark & ICCV'25 & Camera$^*$ & None
& 0.61 & 1.17 & 1.98 & \cellcolor{gray!30}1.25
& 0.01 & 0.22 & 0.85 & \cellcolor{gray!30}0.36 \\
\midrule
\rowcolor{gray!30}
\multicolumn{13}{@{}l}{\raggedright \textit{UNIVERSE Variants}} \\
\midrule
UNIVERSE
& \xmark & -- & Camera$^*$ & None
& \textbf{0.31} & \textbf{0.76} & 1.44 & \cellcolor{gray!30}\textbf{0.83}
& \textbf{0.00} & \textbf{0.03} & \textbf{0.05} & \cellcolor{gray!30}\textbf{0.03}  \\
UNIVERSE w.o. Mask
& \xmark & -- & Camera$^*$ & None
& 0.33 & \textbf{0.76} & \textbf{1.43} & \cellcolor{gray!30}0.84
& \textbf{0.00} & 0.07 & 0.12 & \cellcolor{gray!30}0.06 \\
UNIVERSE w.o. video co-train
& \xmark & -- & Camera$^*$ & None
& 0.48 & 0.94 & 1.65 & \cellcolor{gray!30}1.02
& 0.37 & 0.31 & 0.38 & \cellcolor{gray!30}0.35 \\
UNIVERSE Two-DiT
& \xmark & -- & Camera$^*$ & None
& 0.44 & 0.81 & 1.52 & \cellcolor{gray!30}0.92
& 0.17 & 0.22 & 0.25 & \cellcolor{gray!30}0.21 \\
\bottomrule
\end{tabular}}
\vspace{-0.2cm}
\end{table*}

\section{Additional Zero-Shot Visualization}
\label{sec:supp_visual}

\begin{figure}[htbp]
    \centering
    \includegraphics[width=1.0\linewidth]{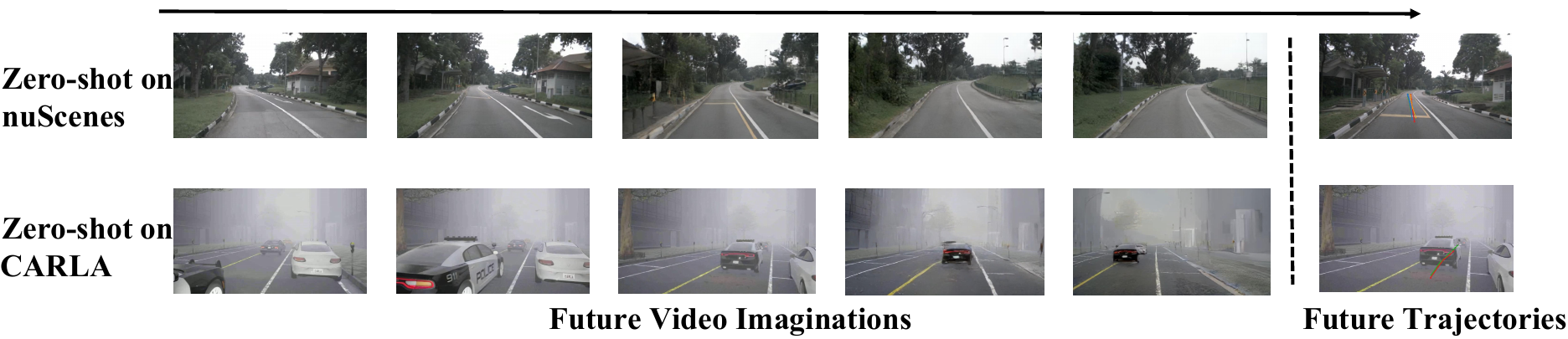}
    \caption{\textbf{Zero-shot qualitative results on nuScenes and Bench2Drive.}
    \textsc{UNIVERSE} is trained on NAVSIM and directly evaluated on unseen target datasets without fine-tuning.
    The generated future videos remain consistent with the predicted ego trajectories across both real-world and simulation-domain scenes.
    For nuScenes, DPVO~\citep{teed2023deep} reconstructions from ground-truth and generated videos further show that the visual rollouts imply ego motion aligned with the reference and predicted trajectories.}
    \label{fig:supp_zero_shot}
    \vspace{-0.25cm}
\end{figure}

Fig.~\ref{fig:supp_zero_shot} provides additional qualitative results on unseen datasets.
The model is trained on NAVSIM and directly evaluated on nuScenes and Bench2Drive without fine-tuning.
For nuScenes, DPVO~\citep{teed2023deep} reconstructs ego trajectories from both ground-truth and generated future videos.
The close alignment between predicted trajectories, generated video evolution, and DPVO-reconstructed motion indicates that the video and trajectory branches learn consistent future dynamics even under domain shift.
For Bench2Drive, the generated future frames and trajectories also remain temporally coherent despite the real-to-simulation visual and dynamics gap.

\begin{figure}[htbp]
    \centering
    \includegraphics[width=1.0\linewidth]{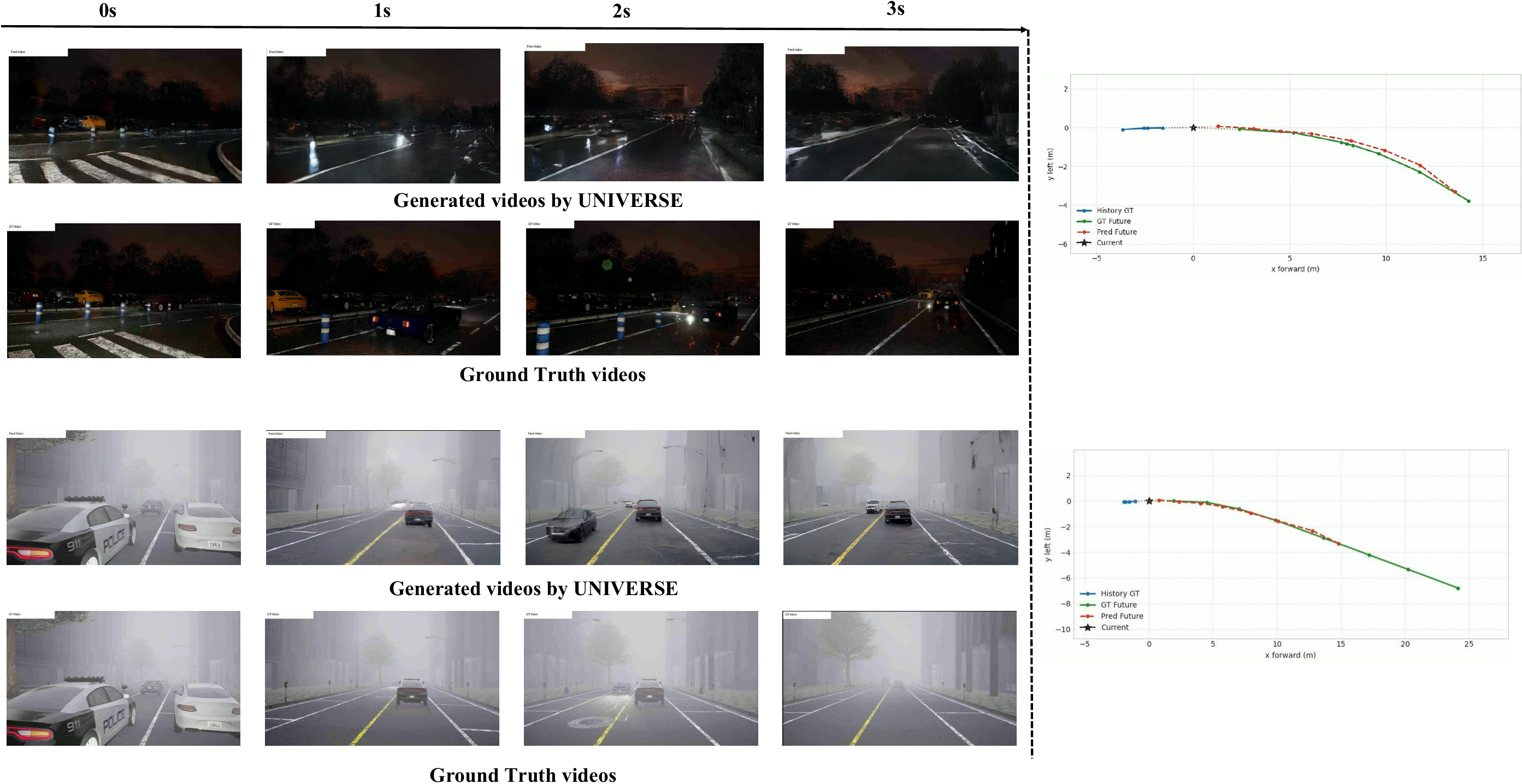}
    \caption{\textbf{Zero-shot video generation and trajectory consistency on Bench2Drive.}
    \textsc{UNIVERSE} is directly evaluated on unseen Bench2Drive scenes without fine-tuning, including challenging nighttime and foggy driving conditions.
    The generated future videos remain visually consistent with the ground-truth rollouts, while the predicted trajectories closely follow the reference ego motion.
    The right-side plots show historical ground-truth motion, ground-truth future motion, predicted future motion, and the current pose. The DPVO~\citep{teed2023deep} consistency protocol is described separately in Sec.~\ref{sec:supp_dpvo}.}
    \label{fig:carla_zero_shot}
    \vspace{-0.25cm}
\end{figure}

\begin{figure}[htbp]
    \centering
    \includegraphics[width=0.96\linewidth]{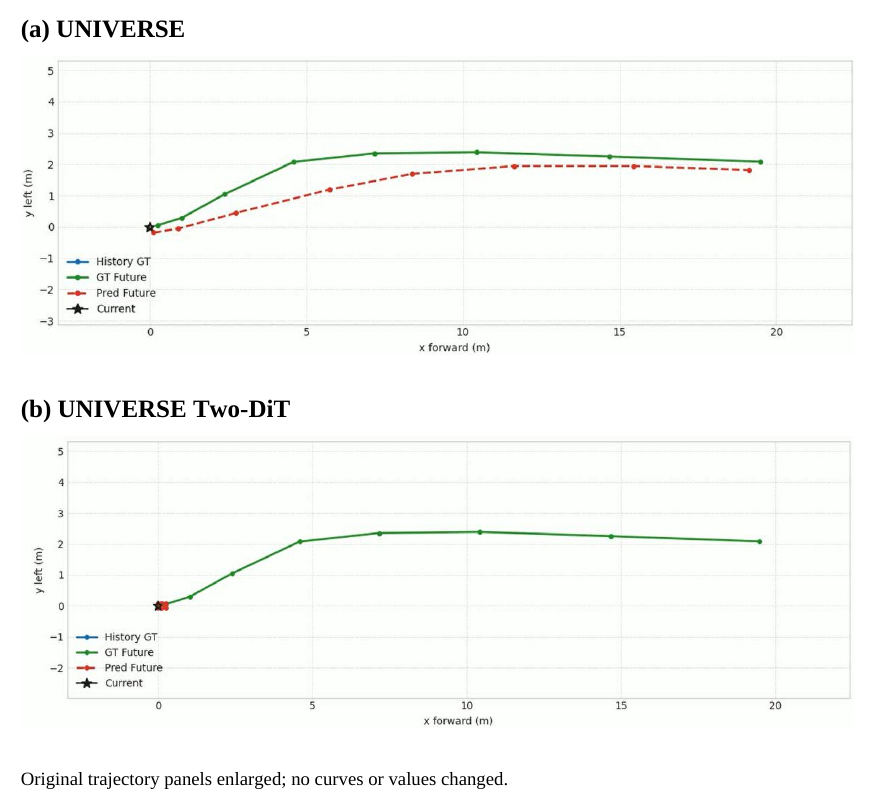}
    \caption{\textbf{Enlarged trajectory comparison for the Bench2Drive example in Fig.~\ref{fig:visual_compare_two_dit}.}
    The panels retain the original curves, axes, and legends. The full model predicts a bypass trajectory, whereas the Two-DiT variant stays close to the current position. No trajectories have been regenerated for this enlargement.}
    \label{fig:supp_trajectory_detail}
\end{figure}
\section{Limitations}
\label{sec:supp_limitations}

Although \textsc{UNIVERSE} achieves strong simulation-based planning and zero-shot generalization, several limitations remain.
First, front-view camera input can miss side or rear-view cues in highly interactive scenes; multi-view extension remains future work.
Second, trajectory-only inference reduces the cost of coupled rollout, but the reported $376$\,ms latency is not a real-vehicle real-time guarantee, and training still relies on a large video-generation backbone.
Third, NAVSIM v1 is non-reactive and the reported cross-domain planning metrics are open-loop. These evaluations do not establish interactive closed-loop performance or real-vehicle safety.
Fourth, backbone comparisons change model family as well as size and are not controlled scaling-law experiments. Aggregate results do not include repeated-run confidence intervals, so small numerical differences should not be interpreted as statistically established gains.
Finally, NAVSIM Train--Navtest gaps and zero-shot performance support improved generalization but do not uniquely identify reliance on source-domain action priors. Future-scene generation and joint rollout are evaluated as model capabilities, not as validated downstream data-augmentation or safety-certification systems.

\section{Licenses for Existing Assets}
\label{sec:supp_licenses}

We use public datasets, benchmarks, models, and evaluation tools according to their official licenses and terms of use.

\noindent\textbf{NAVSIM and OpenScene.}
We use NAVSIM v1 for non-reactive simulation-based planning evaluation. Its code and assets are under Apache-2.0 unless otherwise specified; datasets retain their own licenses.
OpenScene code is released under Apache-2.0, whereas the OpenScene data are based on nuPlan and distributed under Creative Commons Attribution-NonCommercial-ShareAlike and the nuPlan Dataset License Agreement for Non-Commercial Use.
We follow these licenses and use the data for research training and evaluation.

\noindent\textbf{nuScenes.}
We use the official nuScenes validation split for zero-shot planning and video-generation evaluation under its data license and terms of use (CC BY-NC-SA 4.0 for non-commercial research).

\noindent\textbf{Bench2Drive and CARLA.}
We use Bench2Drive for real-to-simulation zero-shot transfer evaluation.
The Bench2Drive dataset page states that its assets and code are released under Apache-2.0 unless specified otherwise.
Bench2Drive is built on CARLA; CARLA-specific code is released under the MIT license, CARLA-specific assets under CC-BY, and third-party dependencies such as Unreal Engine follow their own licenses.
We follow the corresponding licenses and use the benchmark only for research evaluation.

\noindent\textbf{Wan2.2-TI2V-5B.}
The pretrained Wan2.2-TI2V-5B backbone is used under its Apache-2.0 model-card license and release terms.

% \noindent\textbf{Additional video backbones.}
% The backbone comparison uses Cosmos-Predict2.5-2B~\citep{nvidia2025cosmospredict25card} and Cosmos3-Edge (4B)~\citep{nvidia2026cosmos3edgecard}.
% Their official model cards provide the corresponding release terms; the NVIDIA Open Model License and OpenMDW-1.1, respectively, apply to these model releases.
% This comparison does not redistribute their weights.

\noindent\textbf{DPVO.}
We use DPVO only as an evaluation tool for ego-motion reconstruction from ground-truth and generated videos.
It is not used as a training signal.
The official DPVO repository is released under the MIT license.

\noindent\textbf{Baselines.}
Baseline papers and result provenance are cited. Their code, checkpoints, and assets retain the licenses of the original authors.

\section{Broader Impacts}
\label{sec:supp_broader_impacts}

This work studies video-action world models for autonomous-driving planning.
Its potential benefit is improved action generalization under domain shift, while trajectory-only inference reduces the cost of test-time video rollout.
Incorrect trajectories or plausible but inaccurate future videos could lead to unsafe decisions if deployed without independent validation; benchmark collision rates do not certify real-world safety.
Domain and coverage biases may also cause unequal performance across locations, lighting, or road-user interactions.
We recommend use as a research model, compliance with the original data/model licenses, and independent safety checks before any deployment; these are recommendations, not safeguards validated by our experiments.
Training large video-action models also requires substantial compute, motivating future work on efficient training, compression, and distillation.

\clearpage
% Drafted against the supplied manuscript and author responses.
% Public release confirmed by the authors and checked against the repository.
% Authors should confirm the remaining research-process answers before submission, including
% complete reproducibility details, resource accounting, ethics, and asset licenses.
% Standard NeurIPS 2026 questions and guidelines are retained.
\section*{NeurIPS Paper Checklist}
\begin{enumerate}

\Needspace{10\baselineskip}
\item {\bf Claims}
    \item[] Question: Do the main claims made in the abstract and introduction accurately reflect the paper's contributions and scope?
    \item[] Answer: \answerYes{}
    \item[] Justification: Secs.~\ref{sec:method}--\ref{sec:ablation} support the architecture and experimental claims. The text distinguishes shared-parameter foundations from the new dependency design and limits the interpretation of generalization evidence.
    \item[] Guidelines:
    \begin{itemize}
        \item The answer \answerNA{} means that the abstract and introduction do not include the claims made in the paper.
        \item The abstract and/or introduction should clearly state the claims made, including the contributions made in the paper and important assumptions and limitations. A \answerNo{} or \answerNA{} answer to this question will not be perceived well by the reviewers.
        \item The claims made should match theoretical and experimental results, and reflect how much the results can be expected to generalize to other settings.
        \item It is fine to include aspirational goals as motivation as long as it is clear that these goals are not attained by the paper.
    \end{itemize}

\Needspace{10\baselineskip}
\item {\bf Limitations}
    \item[] Question: Does the paper discuss the limitations of the work performed by the authors?
    \item[] Answer: \answerYes{}
    \item[] Justification: Sec.~\ref{sec:supp_limitations} discusses front-view coverage, training/inference cost, evaluation scope, missing statistical uncertainty, and the limits of action-prior and downstream-use interpretations.
    \item[] Guidelines:
    \begin{itemize}
        \item The answer \answerNA{} means that the paper has no limitation while the answer \answerNo{} means that the paper has limitations, but those are not discussed in the paper.
        \item The authors are encouraged to create a separate ``Limitations'' section in their paper.
        \item The paper should point out any strong assumptions and how robust the results are to violations of these assumptions (e.g., independence assumptions, noiseless settings, model well-specification, asymptotic approximations only holding locally). The authors should reflect on how these assumptions might be violated in practice and what the implications would be.
        \item The authors should reflect on the scope of the claims made, e.g., if the approach was only tested on a few datasets or with a few runs. In general, empirical results often depend on implicit assumptions, which should be articulated.
        \item The authors should reflect on the factors that influence the performance of the approach. For example, a facial recognition algorithm may perform poorly when image resolution is low or images are taken in low lighting. Or a speech-to-text system might not be used reliably to provide closed captions for online lectures because it fails to handle technical jargon.
        \item The authors should discuss the computational efficiency of the proposed algorithms and how they scale with dataset size.
        \item If applicable, the authors should discuss possible limitations of their approach to address problems of privacy and fairness.
        \item While the authors might fear that complete honesty about limitations might be used by reviewers as grounds for rejection, a worse outcome might be that reviewers discover limitations that aren't acknowledged in the paper. The authors should use their best judgment and recognize that individual actions in favor of transparency play an important role in developing norms that preserve the integrity of the community. Reviewers will be specifically instructed to not penalize honesty concerning limitations.
    \end{itemize}

\Needspace{10\baselineskip}
\item {\bf Theory assumptions and proofs}
    \item[] Question: For each theoretical result, does the paper provide the full set of assumptions and a complete (and correct) proof?
    \item[] Answer: \answerNA{}
    \item[] Justification: This is an architectural and empirical study; it does not claim a theorem or theoretical performance guarantee.
    \item[] Guidelines:
    \begin{itemize}
        \item The answer \answerNA{} means that the paper does not include theoretical results.
        \item All the theorems, formulas, and proofs in the paper should be numbered and cross-referenced.
        \item All assumptions should be clearly stated or referenced in the statement of any theorems.
        \item The proofs can either appear in the main paper or the supplemental material, but if they appear in the supplemental material, the authors are encouraged to provide a short proof sketch to provide intuition.
        \item Inversely, any informal proof provided in the core of the paper should be complemented by formal proofs provided in appendix or supplemental material.
        \item Theorems and Lemmas that the proof relies upon should be properly referenced.
    \end{itemize}

\Needspace{10\baselineskip}
\item {\bf Experimental result reproducibility}
    \item[] Question: Does the paper fully disclose all the information needed to reproduce the main experimental results of the paper to the extent that it affects the main claims and/or conclusions of the paper (regardless of whether the code and data are provided or not)?
    \item[] Answer: \answerYes{}
    \item[] Justification: Secs.~\ref{sec:method}, \ref{sec:supp_architecture}, and~\ref{sec:supp_protocol} describe the model and evaluation protocol. The publicly released implementation, configurations, and trained checkpoint provide additional resources for reproducing the method; the evaluation protocol is detailed in Sec.~\ref{sec:supp_zero_shot_protocol}.
    \item[] Guidelines:
    \begin{itemize}
        \item The answer \answerNA{} means that the paper does not include experiments.
        \item If the paper includes experiments, a \answerNo{} answer to this question will not be perceived well by the reviewers: Making the paper reproducible is important, regardless of whether the code and data are provided or not.
        \item If the contribution is a dataset and\slash or model, the authors should describe the steps taken to make their results reproducible or verifiable.
        \item Depending on the contribution, reproducibility can be accomplished in various ways. For example, if the contribution is a novel architecture, describing the architecture fully might suffice, or if the contribution is a specific model and empirical evaluation, it may be necessary to either make it possible for others to replicate the model with the same dataset, or provide access to the model. In general. releasing code and data is often one good way to accomplish this, but reproducibility can also be provided via detailed instructions for how to replicate the results, access to a hosted model (e.g., in the case of a large language model), releasing of a model checkpoint, or other means that are appropriate to the research performed.
        \item While NeurIPS does not require releasing code, the conference does require all submissions to provide some reasonable avenue for reproducibility, which may depend on the nature of the contribution. For example\begin{enumerate}\item If the contribution is primarily a new algorithm, the paper should make it clear how to reproduce that algorithm.\item If the contribution is primarily a new model architecture, the paper should describe the architecture clearly and fully.\item If the contribution is a new model (e.g., a large language model), then there should either be a way to access this model for reproducing the results or a way to reproduce the model (e.g., with an open-source dataset or instructions for how to construct the dataset).\item We recognize that reproducibility may be tricky in some cases, in which case authors are welcome to describe the particular way they provide for reproducibility. In the case of closed-source models, it may be that access to the model is limited in some way (e.g., to registered users), but it should be possible for other researchers to have some path to reproducing or verifying the results.\end{enumerate}
    \end{itemize}

\Needspace{10\baselineskip}
\item {\bf Open access to data and code}
    \item[] Question: Does the paper provide open access to the data and code, with sufficient instructions to faithfully reproduce the main experimental results, as described in supplemental material?
    \item[] Answer: \answerYes{}
    \item[] Justification: The implementation, executable configurations, and model checkpoint are publicly released, while the evaluation protocol is detailed in Sec.~\ref{sec:supp_zero_shot_protocol}. Existing benchmark data are obtained from their original providers under the terms summarized in Sec.~\ref{sec:supp_licenses}.
    \item[] Guidelines:
    \begin{itemize}
        \item The answer \answerNA{} means that paper does not include experiments requiring code.
        \item Please see the NeurIPS code and data submission guidelines (\url{https://neurips.cc/public/guides/CodeSubmissionPolicy}) for more details.
        \item While we encourage the release of code and data, we understand that this might not be possible, so \answerNo{} is an acceptable answer. Papers cannot be rejected simply for not including code, unless this is central to the contribution (e.g., for a new open-source benchmark).
        \item The instructions should contain the exact command and environment needed to run to reproduce the results. See the NeurIPS code and data submission guidelines (\url{https://neurips.cc/public/guides/CodeSubmissionPolicy}) for more details.
        \item The authors should provide instructions on data access and preparation, including how to access the raw data, preprocessed data, intermediate data, and generated data, etc.
        \item The authors should provide scripts to reproduce all experimental results for the new proposed method and baselines. If only a subset of experiments are reproducible, they should state which ones are omitted from the script and why.
        \item At submission time, to preserve anonymity, the authors should release anonymized versions (if applicable).
        \item Providing as much information as possible in supplemental material (appended to the paper) is recommended, but including URLs to data and code is permitted.
    \end{itemize}

\Needspace{10\baselineskip}
\item {\bf Experimental setting/details}
    \item[] Question: Does the paper specify all the training and test details (e.g., data splits, hyperparameters, how they were chosen, type of optimizer) necessary to understand the results?
    \item[] Answer: \answerYes{}
    \item[] Justification: The experimental setup reports the optimizer, learning rate, batch schedule, resolution, horizon, and sampling steps. Secs.~\ref{sec:supp_additional_experiments} and~\ref{sec:supp_protocol} document loss sensitivity, inputs, preprocessing, splits, and the open-loop evaluation settings.
    \item[] Guidelines:
    \begin{itemize}
        \item The answer \answerNA{} means that the paper does not include experiments.
        \item The experimental setting should be presented in the core of the paper to a level of detail that is necessary to appreciate the results and make sense of them.
        \item The full details can be provided either with the code, in appendix, or as supplemental material.
    \end{itemize}

\Needspace{10\baselineskip}
\item {\bf Experiment statistical significance}
    \item[] Question: Does the paper report error bars suitably and correctly defined or other appropriate information about the statistical significance of the experiments?
    \item[] Answer: \answerNo{}
    \item[] Justification: The tables report aggregate metrics without repeated-run error bars or confidence intervals. Sec.~\ref{sec:supp_limitations} explicitly cautions against interpreting small differences as statistically established gains.
    \item[] Guidelines:
    \begin{itemize}
        \item The answer \answerNA{} means that the paper does not include experiments.
        \item The authors should answer \answerYes{} if the results are accompanied by error bars, confidence intervals, or statistical significance tests, at least for the experiments that support the main claims of the paper.
        \item The factors of variability that the error bars are capturing should be clearly stated (for example, train/test split, initialization, random drawing of some parameter, or overall run with given experimental conditions).
        \item The method for calculating the error bars should be explained (closed form formula, call to a library function, bootstrap, etc.)
        \item The assumptions made should be given (e.g., Normally distributed errors).
        \item It should be clear whether the error bar is the standard deviation or the standard error of the mean.
        \item It is OK to report 1-sigma error bars, but one should state it. The authors should preferably report a 2-sigma error bar than state that they have a 96\% CI, if the hypothesis of Normality of errors is not verified.
        \item For asymmetric distributions, the authors should be careful not to show in tables or figures symmetric error bars that would yield results that are out of range (e.g., negative error rates).
        \item If error bars are reported in tables or plots, the authors should explain in the text how they were calculated and reference the corresponding figures or tables in the text.
    \end{itemize}

\Needspace{10\baselineskip}
\item {\bf Experiments compute resources}
    \item[] Question: For each experiment, does the paper provide sufficient information on the computer resources (type of compute workers, memory, time of execution) needed to reproduce the experiments?
    \item[] Answer: \answerNo{}
    \item[] Justification: The paper reports NVIDIA H20 training GPUs, the optimization schedule, and measured latency and memory in Table~\ref{tab:efficiency}, but not per-run GPU counts, wall-clock training time, or total project compute.
    \item[] Guidelines:
    \begin{itemize}
        \item The answer \answerNA{} means that the paper does not include experiments.
        \item The paper should indicate the type of compute workers CPU or GPU, internal cluster, or cloud provider, including relevant memory and storage.
        \item The paper should provide the amount of compute required for each of the individual experimental runs as well as estimate the total compute.
        \item The paper should disclose whether the full research project required more compute than the experiments reported in the paper (e.g., preliminary or failed experiments that didn't make it into the paper).
    \end{itemize}

\Needspace{10\baselineskip}
\item {\bf Code of ethics}
    \item[] Question: Does the research conducted in the paper conform, in every respect, with the NeurIPS Code of Ethics \url{https://neurips.cc/public/EthicsGuidelines}?
    \item[] Answer: \answerYes{}
    \item[] Justification: The reported study uses existing research datasets and simulation, credits external assets, and discusses limitations and potential harms in Secs.~\ref{sec:supp_limitations}--\ref{sec:supp_broader_impacts}. No new participant study or real-vehicle deployment is reported.
    \item[] Guidelines:
    \begin{itemize}
        \item The answer \answerNA{} means that the authors have not reviewed the NeurIPS Code of Ethics.
        \item If the authors answer \answerNo, they should explain the special circumstances that require a deviation from the Code of Ethics.
        \item The authors should make sure to preserve anonymity (e.g., if there is a special consideration due to laws or regulations in their jurisdiction).
    \end{itemize}

\Needspace{10\baselineskip}
\item {\bf Broader impacts}
    \item[] Question: Does the paper discuss both potential positive societal impacts and negative societal impacts of the work performed?
    \item[] Answer: \answerYes{}
    \item[] Justification: Sec.~\ref{sec:supp_broader_impacts} discusses generalization and efficiency benefits as well as unsafe predictions, coverage bias, deployment cautions, and computational cost.
    \item[] Guidelines:
    \begin{itemize}
        \item The answer \answerNA{} means that there is no societal impact of the work performed.
        \item If the authors answer \answerNA{} or \answerNo, they should explain why their work has no societal impact or why the paper does not address societal impact.
        \item Examples of negative societal impacts include potential malicious or unintended uses (e.g., disinformation, generating fake profiles, surveillance), fairness considerations (e.g., deployment of technologies that could make decisions that unfairly impact specific groups), privacy considerations, and security considerations.
        \item The conference expects that many papers will be foundational research and not tied to particular applications, let alone deployments. However, if there is a direct path to any negative applications, the authors should point it out. For example, it is legitimate to point out that an improvement in the quality of generative models could be used to generate Deepfakes for disinformation. On the other hand, it is not needed to point out that a generic algorithm for optimizing neural networks could enable people to train models that generate Deepfakes faster.
        \item The authors should consider possible harms that could arise when the technology is being used as intended and functioning correctly, harms that could arise when the technology is being used as intended but gives incorrect results, and harms following from (intentional or unintentional) misuse of the technology.
        \item If there are negative societal impacts, the authors could also discuss possible mitigation strategies (e.g., gated release of models, providing defenses in addition to attacks, mechanisms for monitoring misuse, mechanisms to monitor how a system learns from feedback over time, improving the efficiency and accessibility of ML).
    \end{itemize}

\Needspace{10\baselineskip}
\item {\bf Safeguards}
    \item[] Question: Does the paper describe safeguards that have been put in place for responsible release of data or models that have a high risk for misuse (e.g., pre-trained language models, image generators, or scraped datasets)?
    \item[] Answer: \answerNo{}
    \item[] Justification: The paper discusses research-use and deployment cautions, but does not report implemented or validated release-control safeguards for the model. Recommendations in Sec.~\ref{sec:supp_broader_impacts} are not represented as completed safety measures.
    \item[] Guidelines:
    \begin{itemize}
        \item The answer \answerNA{} means that the paper poses no such risks.
        \item Released models that have a high risk for misuse or dual-use should be released with necessary safeguards to allow for controlled use of the model, for example by requiring that users adhere to usage guidelines or restrictions to access the model or implementing safety filters.
        \item Datasets that have been scraped from the Internet could pose safety risks. The authors should describe how they avoided releasing unsafe images.
        \item We recognize that providing effective safeguards is challenging, and many papers do not require this, but we encourage authors to take this into account and make a best faith effort.
    \end{itemize}

\Needspace{10\baselineskip}
\item {\bf Licenses for existing assets}
    \item[] Question: Are the creators or original owners of assets (e.g., code, data, models), used in the paper, properly credited and are the license and terms of use explicitly mentioned and properly respected?
    \item[] Answer: \answerNo{}
    \item[] Justification: Sec.~\ref{sec:supp_licenses} credits the major assets and states their licenses, including the additional video backbones. A complete per-version inventory of every baseline implementation and auxiliary component is not provided in this version.
    \item[] Guidelines:
    \begin{itemize}
        \item The answer \answerNA{} means that the paper does not use existing assets.
        \item The authors should cite the original paper that produced the code package or dataset.
        \item The authors should state which version of the asset is used and, if possible, include a URL.
        \item The name of the license (e.g., CC-BY 4.0) should be included for each asset.
        \item For scraped data from a particular source (e.g., website), the copyright and terms of service of that source should be provided.
        \item If assets are released, the license, copyright information, and terms of use in the package should be provided. For popular datasets, \url{paperswithcode.com/datasets} has curated licenses for some datasets. Their licensing guide can help determine the license of a dataset.
        \item For existing datasets that are re-packaged, both the original license and the license of the derived asset (if it has changed) should be provided.
        \item If this information is not available online, the authors are encouraged to reach out to the asset's creators.
    \end{itemize}

\Needspace{10\baselineskip}
\item {\bf New assets}
    \item[] Question: Are new assets introduced in the paper well documented and is the documentation provided alongside the assets?
    \item[] Answer: \answerYes{}
    \item[] Justification: The released implementation and UNIVERSE checkpoint are documented with setup and inference instructions. The manuscript specifies model and data settings (Secs.~\ref{sec:method} and~\ref{sec:supp_zero_shot_protocol}); no new dataset is introduced.
    \item[] Guidelines:
    \begin{itemize}
        \item The answer \answerNA{} means that the paper does not release new assets.
        \item Researchers should communicate the details of the dataset\slash code\slash model as part of their submissions via structured templates. This includes details about training, license, limitations, etc.
        \item The paper should discuss whether and how consent was obtained from people whose asset is used.
        \item At submission time, remember to anonymize your assets (if applicable). You can either create an anonymized URL or include an anonymized zip file.
    \end{itemize}

\Needspace{10\baselineskip}
\item {\bf Crowdsourcing and research with human subjects}
    \item[] Question: For crowdsourcing experiments and research with human subjects, does the paper include the full text of instructions given to participants and screenshots, if applicable, as well as details about compensation (if any)?
    \item[] Answer: \answerNA{}
    \item[] Justification: The reported experiments use existing driving datasets and simulation, rather than new crowdsourcing or participant studies.
    \item[] Guidelines:
    \begin{itemize}
        \item The answer \answerNA{} means that the paper does not involve crowdsourcing nor research with human subjects.
        \item Including this information in the supplemental material is fine, but if the main contribution of the paper involves human subjects, then as much detail as possible should be included in the main paper.
        \item According to the NeurIPS Code of Ethics, workers involved in data collection, curation, or other labor should be paid at least the minimum wage in the country of the data collector.
    \end{itemize}

\Needspace{10\baselineskip}
\item {\bf Institutional review board (IRB) approvals or equivalent for research with human subjects}
    \item[] Question: Does the paper describe potential risks incurred by study participants, whether such risks were disclosed to the subjects, and whether Institutional Review Board (IRB) approvals (or an equivalent approval/review based on the requirements of your country or institution) were obtained?
    \item[] Answer: \answerNA{}
    \item[] Justification: No new human-subject experiment or participant recruitment is reported in this work.
    \item[] Guidelines:
    \begin{itemize}
        \item The answer \answerNA{} means that the paper does not involve crowdsourcing nor research with human subjects.
        \item Depending on the country in which research is conducted, IRB approval (or equivalent) may be required for any human subjects research. If you obtained IRB approval, you should clearly state this in the paper.
        \item We recognize that the procedures for this may vary significantly between institutions and locations, and we expect authors to adhere to the NeurIPS Code of Ethics and the guidelines for their institution.
        \item For initial submissions, do not include any information that would break anonymity (if applicable), such as the institution conducting the review.
    \end{itemize}

\Needspace{10\baselineskip}
\item {\bf Declaration of LLM usage}
    \item[] Question: Does the paper describe the usage of LLMs if it is an important, original, or non-standard component of the core methods in this research? Note that if the LLM is used only for writing, editing, or formatting purposes and does \emph{not} impact the core methodology, scientific rigor, or originality of the research, declaration is not required.
    \item[] Answer: \answerNA{}
    \item[] Justification: The reported core method does not use an LLM as an important, original, or non-standard methodological component. The frozen pretrained text encoder used for conditioning is described in Sec.~\ref{sec:unified}.
    \item[] Guidelines:
    \begin{itemize}
        \item The answer \answerNA{} means that the core method development in this research does not involve LLMs as any important, original, or non-standard components.
        \item Please refer to our LLM policy in the NeurIPS handbook for what should or should not be described.
    \end{itemize}

\end{enumerate}

\end{document}